%% file: aaai2026.tex
\documentclass[letterpaper]{article} 
\usepackage{aaai2026}  
\usepackage{times}  
\usepackage{helvet}  
\usepackage{courier}  
\usepackage[hyphens]{url}  
\usepackage{graphicx} 
\usepackage{natbib}  
\usepackage{caption} 
\usepackage{algorithm}
\usepackage{algorithmic}

\usepackage{newfloat}
\usepackage{listings}
\DeclareCaptionStyle{ruled}{labelfont=normalfont,labelsep=colon,strut=off} 
\floatstyle{ruled}
\newfloat{listing}{tb}{lst}{}
\floatname{listing}{Listing}
\title{Raw Data Matters: Enhancing Prompt Tuning by Internal Augmentation on Vision-Language Models}
\author {
    \small
    Haoyang Li\textsuperscript{\rm 1,\rm 2}\thanks{This work has been submitted to the IEEE for possible publication. Copyright may be transferred without notice, after which this version may no longer be accessible.},
    Liang Wang\textsuperscript{\rm 1,\rm 2},
    Chao Wang\textsuperscript{\rm 2}\thanks{Corresponding authors.},
    Siyu Zhou\textsuperscript{\rm 1},
    Jing Jiang\textsuperscript{\rm 1},
    Yan Peng\textsuperscript{\rm 2}\footnotemark[1],
    Guodong Long \textsuperscript{\rm 1}\footnotemark[1]}

\affiliations {
    \textsuperscript{\rm 1}Australian Artificial Intelligence Institute, University of Technology Sydney, Sydney, Australia\\
    \textsuperscript{\rm 2}School of Mechanical Engineering and Automation, Shanghai University, Shanghai, China\\

    haoyang.li-3@student.uts.edu.au, \{cwang, pengyan\}@shu.edu.cn, guodong.long@uts.edu.au
}

\usepackage{bibentry}

\usepackage{graphicx}
\usepackage{multirow}
\usepackage{amsmath}  
\usepackage{amssymb}  

\usepackage {colortbl}
\usepackage{booktabs}

\usepackage{pifont}
\usepackage{bbding}  

\usepackage{cuted}

\begin{document}

\maketitle

\begin{abstract}
For CLIP-based prompt tuning, introducing more data as additional knowledge for enhancing fine-tuning process is proved to be an effective approach. Existing data amplification strategies for prompt tuning typically rely on external knowledge (e.g., large language models or pre-structured knowledge bases), resulting in higher costs for data collection and processing, while generally ignoring further utilization of features in image modality. To address this, we propose \textbf{Aug}mentation-driven \textbf{P}rompt \textbf{T}uning (\texttt{AugPT}), a self-contained distillation-based prompt tuning approach using only internal augmentation on raw dataset to better exploit known features. Specifically, \texttt{AugPT} employs self-supervised augmentation on unlabeled images in the training set, and introduces a novel gating mechanism based on consensus test, reusing the pre-trained prompt tuning backbone model to spontaneously filter noisy samples, further enhancing the quality of augmented views. Extensive experiments validate that \texttt{AugPT} simultaneously enhances model performance and generalization capability without using appended external knowledge. The code of \texttt{AugPT} is available at: \textit{https://github.com/JREion/AugPT}.
\end{abstract}


\section{Introduction}
Bridging images and text, Vision-Language Models (VLMs, represented by CLIP \cite{radford2021clip}) pre-trained on hyper-scale image-text pairs have demonstrated impressive cross-modal alignment and fusion capabilities \cite{li2023blip, alayrac2022flamingo}. To efficiently adapt pretrained VLMs for classification-related downstream tasks, CLIP-based prompt tuning is proposed as a parameter-efficient fine-tuning (PEFT) approach \cite{xing2024peftsurvey}. Freezing all parameters in CLIP encoders, prompt tuning introduces a set of lightweight learnable prompt vectors that replace hard prompts in raw inputs \cite{zhou2022coop, zhou2022cocoop}. As self-adaptive queries, prompt vectors guide the pretrained CLIP toward the target distribution through efficient fine-tuning.

\begin{figure*}[t]
  \centering
  \includegraphics[width=0.9\textwidth]{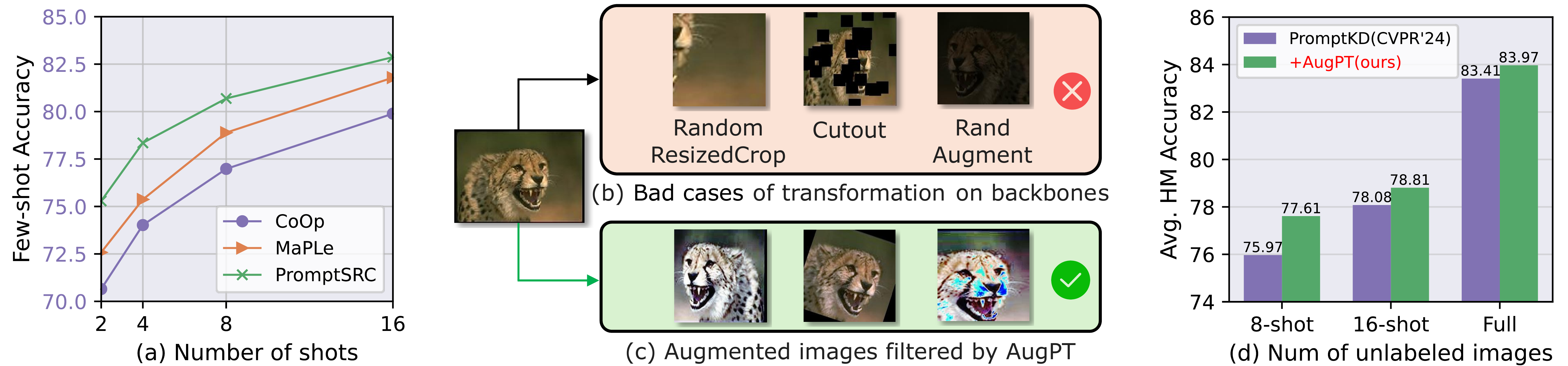}
  \caption{(a) Impact of the amount of data (shots) per class on performance in prompt tuning models \cite{khattak2023maple, yao2023kgcoop, khattak2023promptsrc}; (b) Visualization of bad cases in existing self-supervised augmentation approaches. In comparison, \texttt{AugPT} obtains quality improvement as in (c) by filtering noisy samples, surpassing the current state-of-the-art PromptKD \cite{li2024promptkd} as backbone model in (d) base-to-new tasks in both full and few-shot scenarios over 11 datasets. }
  \label{Figure 1}
\end{figure*}

Given the rareness of annotated image-text pairs in target downstream tasks, extant prompt-tuning strategies \cite{khattak2023maple, yao2023kgcoop, khattak2023promptsrc} are generally carried out under few-shot settings. Therefore, intuitively, introducing more relevant data to the tuning process can enhance recognition performance on target categories (\textbf{base classes}). Empirical evidence from numerous studies \cite{zhou2022coop, khattak2023maple, khattak2023promptsrc} supports this conclusion. As illustrated in Fig.~\ref{Figure 1}(a), with the increase of provided image-text pairs (shots), consistent improvement on fine-tuning performance can be observed on multiple backbone models. This indicates that the prompt tuning process is highly sensitive to the scale of data.

Motivated by this characteristic, numerous approaches explore optimizing prompt tuning at the data level. Current studies primarily focus on introducing \textit{\textbf{external knowledge}} resources to boost base-class performance and generalization on out-of-distribution categories (\textbf{new classes}), containing pre-built hard prompt templates \cite{yao2023kgcoop, yu2023taskres} or structured knowledge bases \cite{kan2023kapt, zhang2024cocole}, and knowledge generated by Large Language Models (LLMs) \cite{wang2024hpt, tian2024argue}. Despite the effectiveness, these strategies require additional data collection and organization costs. In addition, reliance on task-specific external knowledge also limits the generality and transferability for fine-tuning the models. 
Latest research \cite{li2024promptkd, wu2024caspl, li2025dpc} further proposes methods for data amplification utilizing \textit{\textbf{internal knowledge}} of the model. Typically guided by knowledge distillation, these approaches first fine-tune a prompt tuning backbone under few-shot settings \cite{khattak2023promptsrc} as teacher model, then guide it to execute knowledge completion for available unlabeled images for further fine-tuning a student model. Although distillation-based method achieve state-of-the-art (SOTA) in both base-class accuracy and new-class generalization, there are still notable limitations: (1) Performance gains still depend on the expansion of unlabeled image library, which also requires \textit{extra data collection costs}; (2) Extracted image features are only applied for aligning teacher and student, signifying that \textit{raw images are not adequately exploited}.

To address the above limitations, we further propose \textbf{Aug}mentation-driven \textbf{P}rompt \textbf{T}uning (\texttt{AugPT}), a novel distillation-based approach. Building on existing unlabeled images, \texttt{AugPT} does not require additional data or external knowledge, instead focusing on \textit{\textbf{internal augmentation}} to better exploit raw image features. To achieve this, \texttt{AugPT} first introduces an adaptive self-supervised augmentation module that applies self-contained feature transformation to unlabeled images to diversify feature representations. 

It is worth noting that traditional self-supervised augmentation strategies \cite{cubuk2020randaugment, devries2017cutout, mao2022randomresizedcrop} face a key limitation: they lack an error correction mechanism, which may introduce noisy samples that discard key visual features, as illustrated in Fig. 1(b). Although this limitation can be addressed with additional supervision \cite{sohn2020fixmatch}, to maintain parameter efficiency of prompt tuning, we consider strategies that do not require external supervision. As an effective solution, \texttt{AugPT} introduces a novel Consensus Test filtering gate, leveraging a previously overlooked property of pre-tuned prompt tuning models: \textbf{\textit{logit-level consistency reflects semantic similarity across images}} (details are in \textit{Method} section). Concretely, we reuse the frozen teacher model \cite{li2024promptkd} to perform online inference over both raw and augmented images, and adopt Top-1 frequency consistency as consensus to build the view filtering strategy, ensuring that core semantic features of raw image are preserved in augmented views (Fig.~\ref{Figure 1}(c)). Without the need of external modules, this design injects a self-corrective mechanism into augmentation, thereby enhancing cross-modal alignment.

Experiments across 11 datasets reveal that \texttt{AugPT} exceeds PromptKD \cite{li2024promptkd} (current SOTA) as backbone model on base-class performance, new-class generalization, and cross-dataset transfer, without using additional knowledge or external data. Our main contributions are:

\begin{enumerate}
    \item We propose \texttt{AugPT}. To the best of our knowledge, this is the first distillation-based approach that optimizes prompt tuning through internal data augmentation, without introducing external knowledge or additional data.
    \item \texttt{AugPT} incorporates an adaptive self-supervised augmentation strategy and a filtering gate based on Consensus Test, enabling high-quality data amplification without external constraints.
    \item Through experiments, we verify that the raw data in the prompt tuning pipeline can be further exploited through \texttt{AugPT}. We achieve new SOTA across multiple datasets and downstream tasks.
\end{enumerate}

\section{Related Work}  \label{sec2}

\noindent\textbf{Prompt Tuning in VLMs.} Although CLIP \cite{radford2021clip}-based VLMs has remarkable performance on cross-modal tasks, the Transformer-based encoders \cite{dosovitskiy2020vit} with large parameters make it challenging on full fine-tuning for adapting downstream tasks. Therefore, prompt tuning is proposed as a parameter-efficient strategy \cite{xing2024peftsurvey, han2024peftsurveyindpc}. It does not rely on hard prompts (e.g., ``A photo of a [CLASS]''), but introduces a set of learnable vectors as substitutes. During fine-tuning, all parameters of encoders are frozen, and the prompt vectors used as queries are continuously optimized to fit the distribution of target tasks. Existing research explores various formats of prompts, containing text re-encoding \cite{zhou2022coop, zhu2023prograd, tian2024argue}, concatenation with image features \cite{jia2022vpt}, and joint visual-text encoding \cite{khattak2023maple, khattak2023promptsrc, li2025mao, guo2025mmrl}. Since CLIP-based prompt tuning is typically conducted under few-shot scenarios, applying data amplification strategies like \texttt{AugPT} to enhance performance is intuitive, which is supported by the results in Fig.~\ref{Figure 1}(a).

\begin{figure*}[t]
  \centering
  \includegraphics[width=0.9\textwidth]{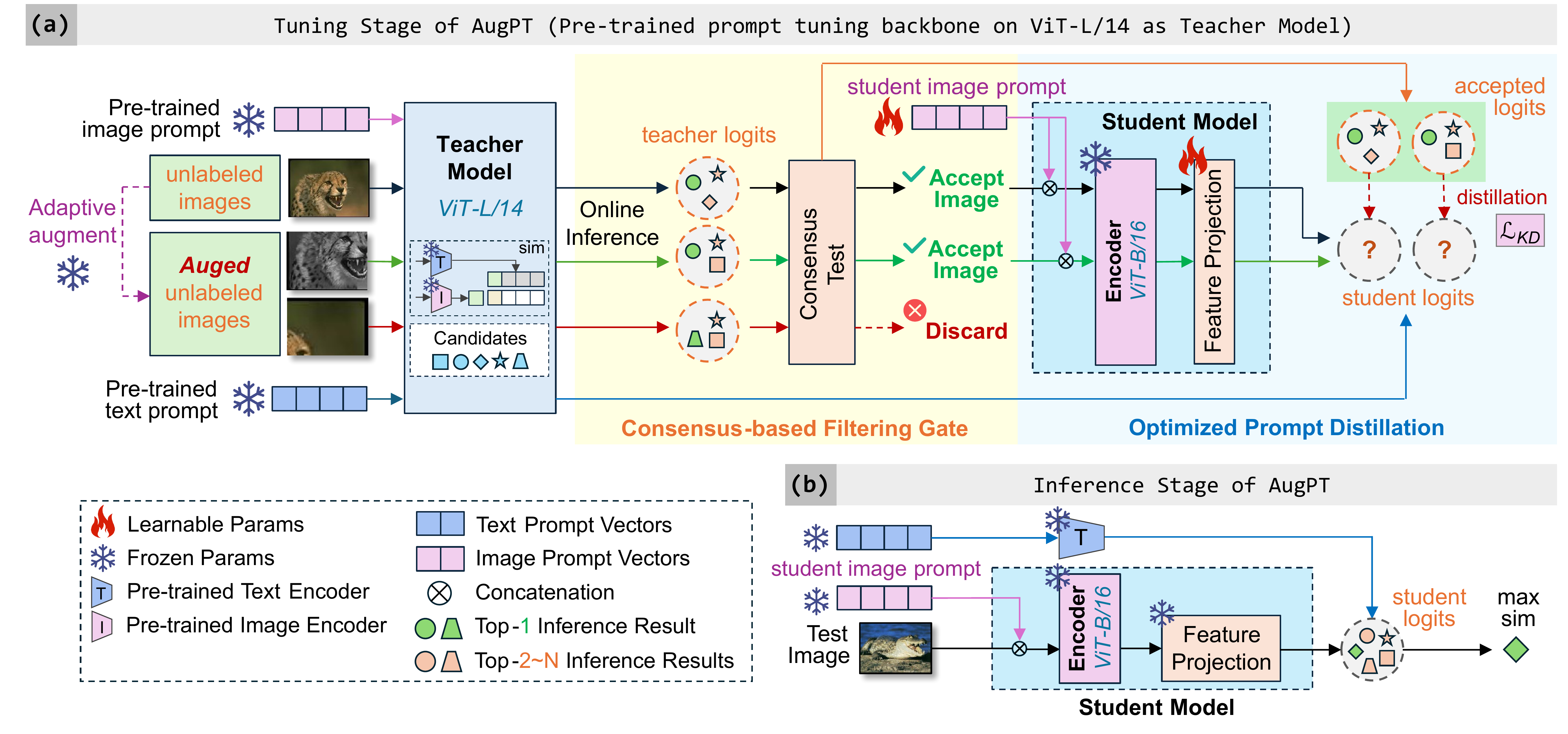}
  \caption{Framework of \texttt{AugPT}. In (a) fine-tuning stage, \texttt{AugPT} applies Adaptive Augmentation to raw unlabeled images $I'$, then reuses the teacher model from distillation-based backbone to discard noisy samples in the augmented image set $\mathcal{D}(I')$ via \textit{Top-1} Consensus-based Filtering Gate, followed by fine-tuning through Optimized Prompt Distillation by fitting the logits of augmented data between teacher and student. Tuned student image prompts $\boldsymbol{P}_{v}^{\text{stu}}$ and projection layer $\text{Proj}(\cdot)$ are applied in (b) inference stage to align student visual features with teacher text embeddings $g^{\text{tea}}({\boldsymbol{P}_{t}^{\text{tea}}})$. }
  \label{Figure 2}
\end{figure*}

\noindent\textbf{Data-Driven Optimization for Prompt Tuning.} Given the limitation of image-text pairs for fine-tuning in CLIP-based prompt tuning, a variety of studies discover data-level optimization strategies. Mainstream approaches consider introducing \textit{\textbf{external knowledge}} to expand data or constrain the optimizing flow, including elaborately designed hard prompt templates as supervision \cite{yao2023kgcoop, khattak2023promptsrc}, or structured databases for appending additional knowledge \cite{kan2023kapt, zhang2024cocole}. More recently, LLMs are employed as external sources to generate descriptions \cite{pratt2023cupl, wang2024hpt, khattak2025protext}, auxiliary attributes \cite{tian2024argue, li2024atprompt, zhu2024awt}, or guide fine-tuning \cite{zheng2024llamp}. Since the above methods demand extra constructing cost for external knowledge bases, the latest works \cite{mistretta2024kdpl, li2024promptkd} propose distillation-based paradigms using \textit{\textbf{internal knowledge + external data}}. Specifically, a teacher model is first fine-tuned on few-shot image-text pairs and then guides a student model by knowledge completion to external unlabeled images. However, further optimizing such models still requires scaling up external data pool, signifying additional collection overhead. In contrast, \texttt{AugPT} proposes an optimized distillation-based strategy that relies only on \textit{\textbf{internal knowledge + internal augmentation}} based on existing raw data.

\section{Proposed Method} \label{sec3}

The framework of \texttt{AugPT} is illustrated in Fig.~\ref{Figure 2}. \texttt{AugPT} adopts the infrastructure of an up-to-date distillation-based prompt tuning model (taking SOTA PromptKD \cite{li2024promptkd} as a paradigm) as its backbone. Existing unlabeled images from the backbone are sequentially processed through an \textit{Adaptive Self-supervised Augmentation} module and a \textit{Consensus-based Filtering Gate}. The pipeline is then optimized for further exploiting latent features within the raw data during fine-tuning.

\subsection{Preliminaries} \label{sec3.1}

\noindent\textbf{Prompt Tuning on CLIP}.  As the foundation model, CLIP is pretrained on $\sim400M$ image-text pairs, employing ViT-based image encoder $f(\cdot)$ and text encoder $g(\cdot)$ to transform images $I$ and texts $T$ to embeddings for cross-modal alignment by feature cosine similarity $sim[\cdot, \cdot]$. To enable rapid adaptation to downstream tasks, prompt tuning freezes all parameters of encoders, while introducing a set of learnable vectors with the size $M \times dim$ to construct text prompts:
\begin{equation}
    \boldsymbol{P}_t=[\mathrm{p}]_{1}[\mathrm{p}]_{2}\ldots[\mathrm{p}]_{M}[CLS]
\end{equation}
where $[CLS]$ denotes the class tokens corresponding to the candidate set $\mathbf{C} = \{ T_i \}_{i=1}^{c}$. Optional image prompts $\boldsymbol{P}_v$ are introduced as a prefix to the image patch tokens, forming visual inputs in the format of $(\boldsymbol{P}_v, I)$. During fine-tuning, the prompt vectors are optimized using cross-entropy loss with a temperature parameter $\tau$, where $h_i$ represents the one-hot label of the object from $\mathbf{C}$:
\begin{equation}
    \mathcal{L}_{\mathrm{CE}}=-\sum_{i} {h}_{i} \log \frac{\exp \left(sim[g({\boldsymbol{P}_t}_y), f(\boldsymbol{P}_v,I)] / \tau\right)}{\sum_{i=1}^{c} \exp \left(sim[g(\boldsymbol{P}_{t_i}), f(\boldsymbol{P}_v,I)]/ \tau\right)}
    \label{Eqn.2}
\end{equation}

\noindent\textbf{Distillation-based Prompt Tuning.}  As the latest prompt tuning optimization strategies, distillation-based approaches first introduce a prompt tuning backbone with more robust encoders, $f^{\text{tea}}(\cdot)$ and $g^{\text{tea}}(\cdot)$ (e.g., PromptSRC with larger ViT-L/14 encoders adopted by PromptKD, details are in \textit{Supplementary Material A.2}), to construct as a \textit{teacher model}. Through the fine-tuning workflow defined in Eqn.~\ref{Eqn.2}, optimized teacher text prompts $\boldsymbol{P}_{t}^{\text{tea}}$ and image prompts $\boldsymbol{P}_{v}^{\text{tea}}$ can be obtained. Next, a \textit{student model} is joined with a standard ViT-B/16 image encoder $f^{\text{stu}}(\cdot)$ and learnable student image prompts $\boldsymbol{P}_{v}^{\text{stu}}$, and the frozen teacher model is applied to predict text-image similarity logits for given unlabeled images $I'$ for knowledge completion. During student fine-tuning, Kullback-Leibler (KL) divergence is introduced to minimize the discrepancy between the teacher logits and the student logits obtained by the interaction of the learnable student visual input  $(\boldsymbol{P}_{v}^{\text{stu}}, I')$ and the frozen teacher text input $\boldsymbol{P}_{t}^{\text{tea}}$, encouraging the distribution of student to align closely with teacher:
\begin{equation}
\begin{aligned}
    \mathcal{L}_{\mathrm{KD}} = \text{KL}(
        & \, \text{sim}\left[g^{\text{tea}}({\boldsymbol{P}_{t}^{\text{tea}}}), f^{\text{tea}}({\boldsymbol{P}_{v}^{\text{tea}}},I')\right], \\
        & \, \text{sim}\left[g^{\text{tea}}({\boldsymbol{P}_{t}^{\text{tea}}}), \text{Proj}\left(f^{\text{stu}}({\boldsymbol{P}_{v}^{\text{stu}}},I')\right)\right]
    )
\end{aligned}
\end{equation}
Herein, $\text{Proj}(\cdot)$ is a learnable MLP-based projection layer, aligning the shape of visual embeddings of student with the text features of teacher model for modal interaction.

\subsection{Adaptive Self-supervised Augmentation} \label{sec3.2}

For the existing raw unlabeled images $I' \in \mathbb{R}^{3 \times H \times W}$ in the prompt tuning backbone, \texttt{AugPT} first performs internal data augmentation to construct a diverse set of views capturing different perspectives and regions of each raw image. This strategy aims to more fully cover visual details, enabling more thorough extraction of latent feature representations in the raw images.

To avoid introducing additional learnable layers or external knowledge, thereby minimizing the cost of data collection and computational overhead, \texttt{AugPT} adopts parameter-free image augmentation based on self-supervised strategy. Replacing the naive Random Resized Crop \cite{mao2022randomresizedcrop} applied in standard prompt tuning, based on the RandAugment \cite{cubuk2020randaugment} methodology, \texttt{AugPT} performs $N$ augmentations on each raw unlabeled image $I'$. For each augmentation, the model samples an operation from a pre-designed set with $Q$ candidates of mainstream transformation policies $\mathcal{Y} = \{\gamma_1, \dots, \gamma_Q\}$ (details are in \textit{Supplementary Material A.3}). Guided by a searching space defined by step size $S$ and amplitude $A$, augmentation operations $\gamma^{(s)} \sim \text{Uniform}(\mathcal{Y})$ are independently and identically sampled at each step $s\in[S]$ to form a policy group $\mathcal{G}$:
\begin{equation}
    \mathcal{G}(I',A, S) =  \Big( \prod_{s=1}^{S} \circ \, ( \gamma^{(s)}, A ) \Big) (I'))
\end{equation}
It is worth noting that original RandAugment uses a pre-searched amplitude $A$ for image transformations, which may limit its generalization ability across datasets with disparate feature distributions. To address this, for the augmentation operation $\gamma^{(s)} \in \mathcal{Y}$ in each step, \texttt{AugPT} correspondingly introduces a uniform distribution $A^{(s)} \sim \text{U}(0, A_{max})$ for dynamically sampling amplitude to better adapt the augmentation operations to various datasets. Following the above workflow, a set $\mathcal{D}(I')$ comprising raw image $I'$ and its $N$ augmented variants can be constructed:
\begin{equation}
    \mathcal{D}(I') = \{ I' \} \cup \{ \mathcal{G}^{n}(I', A^{(s)}, S) \}_{n=1}^N.
\end{equation}

\begin{table*}[t]
\small
\centering
\setlength\tabcolsep{5.5pt}

\begin{tabular}{c|ccc|ccc|ccc|ccc} 
\toprule
\rowcolor{gray!10} {\cellcolor{gray!10}}        & \multicolumn{3}{c|}{\textbf{Average over 11}}    & \multicolumn{3}{c|}{\textbf{ImageNet}}           & \multicolumn{3}{c|}{\textbf{Caltech101}}         & \multicolumn{3}{c}{\textbf{OxfordPets}}           \\
\rowcolor{gray!10} \multirow{-2}{*}{{\cellcolor{gray!10}}\textbf{Method}}
                                          & Base           & New            & H              & Base           & New            & H              & Base           & New            & H              & Base           & New            & H               \\ 
\midrule
CoOp                                      & 82.69          & 63.22          & 71.66          & 76.47          & 67.88          & 71.92          & 98.00          & 89.81          & 93.73          & 93.67          & 95.29          & 94.47           \\
KAPT                                      & 78.42          & 70.52          & 74.26          & 71.10          & 65.20          & 68.02          & 97.10          & 93.53          & 95.28          & 93.13          & 96.53          & 94.80           \\
PromptSRC                                 & 84.26          & 76.10          & 79.97          & 77.60          & 70.73          & 74.01          & 98.10          & 94.03          & 96.02          & 95.33          & 97.30          & 96.30           \\
HPT                                       & 84.32          & 76.86          & 80.42          & 77.95          & 70.74          & 74.17          & 98.37          & 94.98          & 96.65          & 95.78          & 97.65          & 96.71           \\
ArGue                                     & 83.77          & 78.70          & 81.16          & 76.95          & 71.86          & 74.32          & 98.63          & 94.70          & 96.63          & 96.23          & 98.19          & 97.20           \\
LLaMP                                     & 85.13          & 77.73          & 81.26          & 77.99          & 71.27          & 74.48          & 98.45          & 95.85          & 97.13          & 96.31          & 97.74          & 97.02           \\ 
\midrule
PromptKD                                  & 86.91          & 80.17          & 83.41          & 80.82          & 74.66          & 77.62          & 99.03          & 96.62          & 97.81          & 96.17          & 98.21          & 97.18           \\
\rowcolor{gray!20}
\textbf{+AugPT}                                    & \textbf{87.31} & \textbf{80.87} & \textbf{83.97} & \textbf{81.13} & \textbf{74.81} & \textbf{77.84} & \textbf{99.23} & \textbf{96.72} & \textbf{97.96} & \textbf{96.76} & \textbf{98.32} & \textbf{97.53}  \\

\midrule
\midrule

\rowcolor{gray!10} {\cellcolor{gray!10}} & \multicolumn{3}{c|}{\textbf{StanfordCars}}       & \multicolumn{3}{c|}{\textbf{Flowers102}}         & \multicolumn{3}{c|}{\textbf{Food101}}            & \multicolumn{3}{c}{\textbf{FGVCAircraft}}         \\
\rowcolor{gray!10} \multirow{-2}{*}{{\cellcolor{gray!10}}\textbf{Method}}
                                          & Base           & New            & H              & Base           & New            & H              & Base           & New            & H              & Base           & New            & H               \\ 
\midrule
CoOp                                      & 78.12          & 60.40          & 68.13          & 97.60          & 59.67          & 74.06          & 88.33          & 82.26          & 85.19          & 40.44          & 22.30          & 28.75           \\
KAPT                                     & 69.47          & 66.20          & 67.79          & 95.00          & 71.20          & 81.40          & 86.13          & 87.06          & 86.59          & 29.67          & 28.73          & 29.19           \\
PromptSRC                                & 78.27          & 74.97          & 76.58          & 98.07          & 76.50          & 85.95          & 90.67          & 91.53          & 91.10          & 42.73          & 37.87          & 40.15           \\
HPT                                     & 76.95          & 74.23          & 75.57          & 98.17          & 78.37          & 87.16          & 90.46          & 91.57          & 91.01          & 42.68          & 38.13          & 40.28           \\
ArGue                                    & 75.06          & 74.18          & 74.62          & 98.62          & 77.96          & 87.08          & 91.42          & 92.40          & 91.91          & 41.29          & 38.80          & 40.01           \\
LLaMP                                    & 81.56          & 74.54          & 77.89          & 97.82          & 77.40          & 86.42          & 91.05          & 91.93          & 91.49          & 47.30          & 37.61          & 41.90           \\ 
\midrule
PromptKD                                 & 82.61          & 84.10          & 83.35          & 99.05          & 81.99          & 89.72          & 92.68          & 93.83          & 93.25          & 49.10          & 40.91          & 44.63           \\
\rowcolor{gray!20}
\textbf{+AugPT}                                    & \textbf{84.31} & \textbf{84.35} & \textbf{84.33} & \textbf{99.53} & \textbf{82.48} & \textbf{90.21} & \textbf{92.86} & \textbf{94.04} & \textbf{93.45} & \textbf{49.28} & \textbf{42.59} & \textbf{45.69}  \\

\midrule
\midrule

\rowcolor{gray!10} {\cellcolor{gray!10}} & \multicolumn{3}{c|}{\textbf{SUN397}}             & \multicolumn{3}{c|}{\textbf{DTD}}                & \multicolumn{3}{c|}{\textbf{EuroSAT}}            & \multicolumn{3}{c}{\textbf{UCF101}}               \\
\rowcolor{gray!10} \multirow{-2}{*}{{\cellcolor{gray!10}}\textbf{Method}}
                                          & Base           & New            & H              & Base           & New            & H              & Base           & New            & H              & Base           & New            & H               \\ 
\midrule
CoOp                                     & 80.60          & 65.89          & 72.51          & 79.44          & 41.18          & 54.24          & 92.19          & 54.74          & 68.69          & 84.69          & 56.05          & 67.46           \\
KAPT                                     & 79.40          & 74.33          & 76.78          & 75.97          & 58.30          & 65.97          & 84.80          & 67.57          & 75.21          & 80.83          & 67.10          & 73.33           \\
PromptSRC                                & 82.67          & 78.47          & 80.52          & 83.37          & 62.97          & 71.75          & 92.90          & 73.90          & 82.32          & 87.10          & 78.80          & 82.74           \\
HPT                                      & 82.57          & 79.26          & 80.88          & 83.84          & 63.33          & 72.16          & 94.24          & 77.12          & 84.82          & 86.52          & 80.06          & 83.16           \\
ArGue                                    & 81.89          & 80.48          & 81.18          & 80.33          & 67.03          & 73.08          & 95.10          & 90.68          & 92.84          & 86.00          & 79.43          & 82.58           \\
LLaMP                                    & 83.41          & 79.90          & 81.62          & 83.49          & 64.49          & 72.77          & 91.93          & 83.66          & 87.60          & 87.13          & 80.66          & 83.77           \\ 
\midrule
PromptKD                                 & 83.49          & 81.26          & 82.36          & 86.34          & 68.48          & 76.38          & \textbf{97.10} & 80.85          & 88.23          & \textbf{89.66} & 80.96          & 85.09           \\
\rowcolor{gray!20}
\textbf{+AugPT}                                    & \textbf{83.86} & \textbf{81.66} & \textbf{82.75} & \textbf{86.81} & \textbf{71.26} & \textbf{78.27} & 97.06          & \textbf{81.79} & \textbf{88.77} & 89.56          & \textbf{81.56} & \textbf{85.37}  \\

\bottomrule
\end{tabular}

    \caption{Base-to-new generalization performance (\%) of baselines and our \texttt{AugPT} on 11 datasets. }
    \label{tab1}

\end{table*}

\subsection{Consensus-based Filtering Gate} \label{sec3.3}

Although self-supervised augmentation is simple and efficient, it lacks error self-correction capability and may introduce noisy samples that discard key semantic (Fig.~\ref{Figure 1}(b)). To ensure data quality, further evaluation and filtering are necessary. As the motivation, we observe an overlooked property of prompt tuning models: in inference stage, \textbf{\textit{logit-level consistency reflects semantic similarity across images}}. In other words, augmented views that share distribution-consistent logits tend to exhibit similar semantic, meaning that such view set can be regarded as positive samples that preserve subject semantics.

We attribute this property that: for tuned teacher model, the visual prompts appended to the image are re-frozen, resulting in a deterministic mapping from input image to embedding space, guided by the fine-tuned vision-language alignment as prior. As a consequence, the logits obtained from the interaction between augmented image features and the fixed set of textual features implicitly reflect the high-level visual semantics preserved in the view. \textit{Supplementary Material B.4} gives a theoretical proof of this formulation.

Based on this property, \texttt{AugPT} introduces a Filtering Gate mechanism, reusing the learned prompt tuning backbone as a prior, and performs Argmax over frequency counts of image-to-text similarity logits across the augmented views as consensus, eliminating noisy samples that fail to retain the core semantics through consistency.

Specifically, for $\mathcal{D}(I')$ obtained through Adaptive Augmentation module, \texttt{AugPT} first reuses the teacher model from distillation-based backbone (details are in \textit{Supplementary Material A.2}) to perform online inference on all members $I'_j \in \mathcal{D}(I')$ to acquire similarity logits $\boldsymbol{\ell}_j \in \mathbb{R}^{1 \times c}$ between each image $I'_j$ and textual candidates $\mathbf{C} = \{ T_i \}_{i=1}^{c}$:\begin{equation} \label{eqn6}
    \boldsymbol{\ell}_j = sim[g^{\text{tea}}({\boldsymbol{P}_{t}^{\text{tea}}}), f^{\text{tea}}({\boldsymbol{P}_{v}^{\text{tea}}},I'_j)], \quad j = 1, \dots, N+1
\end{equation}

Subsequently, \texttt{AugPT} determines the index corresponding to the maximum logit for each image, and organizes the resulting top-1 predictions into a sequence $\mathbf{u}$:
\begin{equation} \label{eqn7}
    \mathbf{u} = (\theta_1, \theta_2, \dots, \theta_{N+1}), \quad \theta_j = \underset{k \in \{1, \dots, c\}}{\arg\max} \ \boldsymbol{\ell}_j^{(k)}
\end{equation}
where $\theta_j$ is the predicted category index for image $I'_j$. 

Next, to identify the consensus category that best represents the overall distribution of Top-1 predictions, \texttt{AugPT} performs Argmax over frequency counting to the obtained $\mathbf{u}$. The index with the highest occurrence is accepted as the consensus $\theta^\star$ for filtering $\mathcal{D}(I')$:
\begin{equation}
    \theta^\star = \underset{k \in \{1, \dots, c\}}{\arg\max} \ \sum_{j=1}^{N+1} \mathbb{I}(\theta_j = k)
\end{equation}
where $\mathbb{I}(\cdot)$ denotes the indicator function that returns 1 if the condition is satisfied and 0 otherwise.

Finally, the consensus category $\theta^\star$ is applied as the supervision of the Filtering Gate. Only the images with Top-1 predicted categories aligning with $\theta^\star$ are accepted for yielding a refined subset $\mathcal{I}_{acc}$:
\begin{equation}
    \mathcal{I}_{acc} = \{ \mathcal{D}(I') \mid \theta_j = \theta^\star \}
\end{equation}

It is worth noting that the Filtering Gate also applies to raw image $I'$, which would be discarded if it fails to meet the consensus criterion. This strategy allows \texttt{AugPT} to autonomously identify and eliminate noisy samples potentially introduced by the prompt tuning backbone (as shown in Fig.~\ref{Figure 1}(b)-1), thereby improving the quality of passing data for fine-tuning \texttt{AugPT}.

\subsection{Optimized Prompt Distillation} \label{sec3.4}

Based on the filtered image set $\mathcal{I}_{\text{acc}}$ and the consensus category $\theta^\star$, \texttt{AugPT} executes further optimization following the fine-tuning workflow of distillation-based prompt tuning backbone models \cite{li2024promptkd}. Specifically, under the constraint of $\theta^\star$, \texttt{AugPT} samples the teacher logits $\hat{\mathcal{\ell}}^{\text{tea}}_p$ corresponding to each accepted image $\hat{I}_p \in \mathcal{I}_{\text{acc}}$ to organize a set $\mathcal{T}^{\text{tea}}_{\text{acc}}$ in a one-to-one correspondence manner:
\begin{equation}
    \mathcal{T}^{\ tea}_{acc} = \{ \hat {\mathcal{\ell}}^{\ tea}_p \mid \theta_p = \theta^\star \}, \quad p=1, 2, \dots, |\mathcal{I}_{acc}|
\end{equation}
Next, \texttt{AugPT} constructs a student model identical to the backbone (which a frozen CLIP-ViT-B/16 image encoder $f^{\text{stu}}(\cdot)$ and learnable student image prompts $\boldsymbol{P}_{v}^{\text{stu}}$). Similarly, KL divergence is utilized to align the distributions between teacher logits $\hat{\mathcal{\ell}}^{\text{tea}}_p$ from the filtered set and student logits $\hat{\mathcal{\ell}}^{\text{stu}}_p=sim[g^{\text{tea}}({\boldsymbol{P}_{t}^{\text{tea}}}), \text{Proj}(f^{\text{stu}}({\boldsymbol{P}_{v}^{\text{stu}}},\hat{I_p}))]$, which are obtained by interacting the learnable student visual inputs $(\boldsymbol{P}_{v}^{\text{stu}}, \hat{I}_p)$ with the frozen teacher text prompts $\boldsymbol{P}_{t}^{\text{tea}}$. Overall, $\boldsymbol{P}_{v}^{\text{stu}}$ is continuously optimized by:
\begin{equation}
    \mathcal{L}_{\mathrm{KD}}={\frac{1}{|\mathcal{I}_{acc}|}}\sum_{p=1}^{|\mathcal{I}_{acc}|} \text{KL}\left(\hat {\mathcal{\ell}}^{\ tea}_p, \hat{\mathcal{\ell}}^{\ stu}_p \right)
\end{equation}
Herein, $\text{Proj}(\cdot)$ refers to an MLP-based learnable projection layer that is identical to the backbone model \cite{li2024promptkd}, which aligns the student's ViT-B/16-based visual features ($dim=512$) with the teacher's ViT-L/14-based text features ($dim=768$). Aforementioned fine-tuning process promotes \texttt{AugPT} to optimize cross-modal alignment across all filtered augmented views, more effectively exploiting the latent features in the raw data.

\begin{table*}[t]

\small
\centering
\setlength\tabcolsep{3pt}

\begin{tabular}{c|cc|ccccccccccc|c} 
\toprule
\rowcolor{gray!10} \textbf{Shot} & \multicolumn{2}{c|}{\textbf{Method}}                                     & ImgNet                                             & Cal101                                             & Pets                                               & Cars                                               & Flower                                             & Food                                               & Aircraft                                           & SUN                                                & DTD                                                & SAT                                                & UCF                                                & \textbf{Avg.}                                              \\ 
\midrule
\multirow{7}{*}{\textbf{8-shot}}                & \multirow{3}{*}{PromptKD}        & Base                                  & 75.51                                              & 97.55                                              & 91.39                                              & 77.34                                              & 96.39                                              & 86.80                                              & 39.14                                              & 81.28                                              & 74.65                                              & 62.05                                              & 84.23                                              & 78.76                                             \\
                                                &                                  & New                                   & 69.63                                              & 94.76                                              & 95.64                                              & 79.70                                              & 78.30                                              & 89.17                                              & 35.69                                              & 78.84                                              & 61.23                                              & 47.21                                              & 76.96                                              & 73.38                                             \\
                                                &                                  & H                                     & 72.45                                              & 96.13                                              & 93.47                                              & 78.50                                              & 86.41                                              & 87.97                                              & 37.34                                              & 80.04                                              & 67.28                                              & 53.62                                              & 80.43                                              & 75.97                                             \\ 
\cmidrule{2-15}
                                                & \multirow{3}{*}{\textbf{+AugPT}} & Base                                  & 75.88                                                & 98.39                                              & 92.72                                              & 79.31                                              & 97.34                                              & 86.95                                              & 42.08                                              & 81.30                                              & 78.47                                              & 79.55                                              & 84.85                                              & 81.53                                               \\
                                                &                                  & New                                   & 69.42                                                & 94.87                                              & 95.35                                              & 79.67                                              & 79.43                                              & 89.34                                              & 36.89                                              & 79.26                                              & 62.08                                              & 50.33                                              & 77.77                                              & 74.04                                               \\
                                                &                                  & {\cellcolor{gray!20}}H & {\cellcolor{gray!20}}\textbf{72.51}   & {\cellcolor{gray!20}}\textbf{96.60} & {\cellcolor{gray!20}}\textbf{94.02} & {\cellcolor{gray!20}}\textbf{79.49} & {\cellcolor{gray!20}}\textbf{87.48} & {\cellcolor{gray!20}}\textbf{88.13} & {\cellcolor{gray!20}}\textbf{39.31} & {\cellcolor{gray!20}}\textbf{80.27} & {\cellcolor{gray!20}}\textbf{69.32} & {\cellcolor{gray!20}}\textbf{61.65} & {\cellcolor{gray!20}}\textbf{81.16} & {\cellcolor{gray!20}}\textbf{77.60}  \\

                                                & \multicolumn{2}{c|}{$\Delta$}                                                   & +0.06                                                & +0.46                                              & +0.55                                             & +0.99                                              & +1.07                                              & +0.16                                              & +1.98                                              & +0.23                                              & +2.04                                              & +8.03                                              & +0.72                                              & +1.63                                               \\ 
\midrule
\midrule

\multirow{7}{*}{\textbf{16-shot}}               & \multirow{3}{*}{PromptKD}        & Base                                  & 77.50                                              & 98.39                                              & 94.84                                              & 79.81                                              & 97.81                                              & 88.97                                              & 44.60                                              & 82.70                                              & 81.48                                              & 77.02                                              & 87.18                                              & 82.75                                             \\
                                                &                                  & New                                   & 70.98                                              & 95.85                                              & 97.09                                              & 82.00                                              & 81.35                                              & 90.69                                              & 38.81                                              & 80.37                                              & 69.08                                              & 67.54                                              & 78.64                                              & 77.49                                             \\
                                                &                                  & H                                     & 74.10                                              & 97.10                                              & 95.95                                              & 80.89                                              & \textbf{88.82}                                              & 89.82                                              & 41.50                                              & 81.52                                              & 74.77                                              & 71.97                                              & 82.69                                              & 80.04                                             \\ 
\cmidrule{2-15}
                                                & \multirow{3}{*}{\textbf{+AugPT}} & Base                                  & 77.80                                              & 98.97                                              & 96.23                                              & 81.63                                              & 98.48                                                & 89.33                                              & 44.66                                              & 83.17                                              & 82.87                                              & 82.21                                              & 88.11                                              & 83.95                                               \\
                                                &                                  & New                                   & 71.12                                              & 96.62                                              & 97.32                                              & 82.37                                              & 80.78                                                & 91.26                                              & 39.77                                              & 80.98                                              & 69.20                                              & 74.28                                              & 80.80                                              & 78.59                                               \\
                                                &                                  & {\cellcolor{gray!20}}H & {\cellcolor{gray!20}}\textbf{74.31} & {\cellcolor{gray!20}}\textbf{97.78} & {\cellcolor{gray!20}}\textbf{96.77} & {\cellcolor{gray!20}}\textbf{82.00} & {\cellcolor{gray!20}}88.76   & {\cellcolor{gray!20}}\textbf{90.28} & {\cellcolor{gray!20}}\textbf{42.07} & {\cellcolor{gray!20}}\textbf{82.06} & {\cellcolor{gray!20}}\textbf{75.42} & {\cellcolor{gray!20}}\textbf{78.04} & {\cellcolor{gray!20}}\textbf{84.30} & {\cellcolor{gray!20}}\textbf{81.18}  \\
                                                & \multicolumn{2}{c|}{$\Delta$}                                                   & +0.21                                              & +0.68                                              & +0.82                                              & +1.11                                              & -0.07                                                & +0.46                                              & +0.57                                              & +0.54                                              & +0.65                                              & +6.07                                              & +1.61                                              & +1.15                                               \\
\bottomrule
\end{tabular}
    \caption{Base-to-new generalization (\%) of PromptKD backbone and \texttt{AugPT} on 11 datasets with \textbf{few-shot unlabeled images}.}
    \label{tab2}
\end{table*}

\begin{table*}[t]

\small
\centering

\begin{tabular}{c|cccccccccc|c} 
\toprule
\rowcolor{gray!10} {\cellcolor{gray!10}}                                   & \multicolumn{10}{c|}{\textbf{Target Datasets}}                                                                                                                          & {\cellcolor{gray!10}}                                 \\
\rowcolor{gray!10} \multirow{-2}{*}{{\cellcolor{gray!10}}\textbf{Method}} & Cal101         & Pets           & Cars           & Flower         & Food           & Aircraft       & SUN            & DTD            & SAT            & UCF            & \multirow{-2}{*}{{\cellcolor{gray!10}}\textbf{Avg.}}  \\ 
\midrule
CoOp                                                                                              & 93.91          & 89.97          & 65.56          & 67.88          & 85.86          & 22.11          & 66.92          & 42.55          & 47.77          & 67.30          & 64.98                                                             \\
KAPT                                                                                              & 89.63          & 87.60          & 60.73          & 74.17          & 78.07          & 22.13          & 64.50          & 50.93          & 46.50          & 65.90          & 64.02                                                             \\
PromptSRC                                                                                         & 93.43          & 89.92          & 65.95          & 71.05          & 86.21          & 24.03          & 67.63          & 46.22          & 42.59          & 69.39          & 65.64                                                             \\
HPT                                                                                               & 94.20          & 92.63          & 66.33          & 74.84          & 86.21          & 25.68          & 68.75          & 50.87          & 47.36          & 70.50          & 67.74                                                             \\ 
\midrule
PromptKD                                                                                          & 93.23          & \textbf{91.77} & 73.68          & 74.83          & \textbf{89.01} & 27.18          & 67.53          & 53.31          & \textbf{68.31} & 74.65          & 71.35                                                             \\
\rowcolor{gray!20}
\textbf{+AugPT}                                                                                             & \textbf{94.20} & 91.74          & \textbf{74.98} & \textbf{74.87} & 88.94          & \textbf{27.24} & \textbf{68.54} & \textbf{54.43} & 67.22          & \textbf{75.42} & \textbf{71.76}                                                    \\
\bottomrule
\end{tabular}
    \caption{Cross-dataset generalization (\%) of baselines and \texttt{AugPT} on ImageNet as source and 10 other datasets as targets.}
    \label{tab3}
\end{table*}

\section{Experiments} \label{sec4}

\subsection{Experimental Setup} \label{sec4.1}

\noindent \textbf{Datasets.} Following the paradigm of benchmark settings in mainstream prompt tuning models \cite{zhou2022coop, khattak2023promptsrc, li2024promptkd}, \texttt{AugPT} apply 11 recognition-related datasets with diverse domain distributions of data for evaluation on downstream tasks, containing ImageNet \cite{deng2009imagenet}, Caltech101 \cite{fei2004caltech}, OxfordPets \cite{parkhi2012pets}, StanfordCars \cite{krause2013cars}, Flowers102 \cite{nilsback2008flowers}, Food101 \cite{bossard2014food}, FGVCAircraft \cite{maji2013aircraft}, SUN397 \cite{xiao2010sun}, DTD \cite{cimpoi2014dtd}, EuroSAT \cite{helber2019eurosat} and UCF101 \cite{soomro2012ucf101}. 

\noindent \textbf{Baselines.} We select 7 representative \textbf{data-driven} prompt tuning models as baselines, including foundational CoOp \cite{zhou2022coop}, PromptSRC \cite{khattak2023promptsrc}, KAPT \cite{kan2023kapt} that links structured knowledge bases, and typical LLM-guided methods: generating descriptive texts (HPT \cite{wang2024hpt}), attribute augmentation (ArGue \cite{tian2024argue}), and fine-tuning optimization (LLaMP \cite{zheng2024llamp}). Beyond these, we make a detailed comparison with PromptKD \cite{li2024promptkd} (current SOTA), the distillation-based backbone model of \texttt{AugPT}, to quantify the improvement across various tasks. We compare more baselines in \textit{Supplementary Material B.1}. Additionally, to further evaluate the adaptability of \texttt{AugPT}, in \textit{Supplementary Material B.5}, we introduce UPL \cite{zang2022upt} to prove that \texttt{AugPT} is a plug-and-play approach for distillation-based prompt tuning frameworks.

\noindent \textbf{Implementation Details.} We initialize backbone model of \texttt{AugPT} using the optimal parameters obtained through prior exploration by PromptKD, including the pre-tuned teacher model with ViT-L/14 encoders, as well as a student model with a ViT-B/16 encoder $f^{\text{stu}}(\cdot)$ and learnable image prompts $\boldsymbol{P}_{v}^{\text{stu}}$. Identical to PromptKD, $\text{Proj}()$ layer consists of 2 layers. For internal data augmentation, based on ablations, we set the number of augmented images to $N = 5$ and the augmenting step to $S = 2$. Further details and hyperparameters are enumerated in \textit{Supplementary Material A}.

\subsection{Experimental Results of Downstream Tasks} \label{sec4.2}

\paragraph{Base-to-New Generalization.} Inheriting the setup of baselines, classes in each dataset are evenly split into base and new. Consistent with PromptKD, \texttt{AugPT} reuses the teacher model tuned on base classes, and subsequently fine-tunes the student model utilizing all unlabeled images from the training split, while the association between images and new classes is ensured to be completely unknown. Importantly, the teacher and student models must adopt identical class splits to prevent data leakage. Finally, inference is performed on base and new classes to evaluate both accuracy and generalization, where $H$ denotes the Harmonic Mean (HM) between base and new. As in Tab.~\ref{tab1}, \texttt{AugPT} surpasses all external knowledge-driven baselines across all 11 datasets in both base-class accuracy and new-class generalization. Furthermore, \textit{without introducing additional external data}, \texttt{AugPT} outperforms PromptKD backbone overall. Above results demonstrate that the internal data augmentation of \texttt{AugPT} enables more comprehensive learning of raw image features, achieving stronger image-text alignment. Error bar analysis is contained in \textit{Supplementary Material B.2}.

\paragraph{Base-to-New Generalization on Few-shot Unlabeled Images.} Considering the additional cost of collecting unlabeled images in realistic scenes, we introduce a more challenging \textit{Few-shot Base-to-New} task for distillation-based prompt tuning models (PromptKD \& \texttt{AugPT}) to evaluate under data scarcity conditions. Instead of using all unlabeled images, this task only samples few-shot unlabeled images per class to measure base-class fine-tuning and new-class generalization. As shown in Tab.~\ref{tab2}, \texttt{AugPT} still outperforms PromptKD in HM performance across both 8-shot and 16-shot scenarios. Notably, the enhancements are more significant than full-dataset setting, particularly on datasets with extremely limited images (e.g., EuroSAT with $<$200 samples). These results suggest that the strategy of \texttt{AugPT} that exploits internal image features significantly enhances the model's ability to fit target distributions in data-scarce scenarios, highlighting its practical value.

\paragraph{Cross-Dataset Transfer.} Following the \textit{transductive} zero-shot learning setting of PromptKD, \texttt{AugPT} first fine-tunes the teacher model on ImageNet as source data. After this, the student model is fine-tuned using all unlabeled images from other 10 datasets as target (following zero-shot setting, image-text relationships are entirely unknown in target data). As reported in Tab.~\ref{tab3}, \texttt{AugPT} still achieves performance no lower than PromptKD backbone across most datasets, demonstrating strong robustness and reliable generalization in fully unsupervised target domains. Given the distributional differences between source and target domains, these results also help mitigate concerns that internal augmentation may amplify \textit{biases} from the teacher model.

\begin{table}[t]

    \small
    \centering
\setlength{\tabcolsep}{6pt}

    \begin{tabular}{cc|ccc|c} 
    \toprule
    \rowcolor{gray!10} \multicolumn{2}{c|}{\textbf{Sub-modules}} & \multicolumn{3}{c|}{\textbf{Average Performance}} & {\cellcolor{gray!10}}                                                      \\
    \rowcolor{gray!10} ASA & CFG                                 & Base  & New   & H                    & \multirow{-2}{*}{{\cellcolor{gray!10}}\textbf{$\Delta$}}  \\ 
    \midrule
                               \ding{55}     &  \ding{55}                                   & 72.68 & 64.50 & 68.35                &                                                                                     \\
    \ding{51}                               &  \ding{55}                                   & 72.96 & 65.97 & 69.29                & +0.94                                                                               \\
    \rowcolor{gray!20}
    \ding{51}                               & \ding{51}                                   & \textbf{73.47} & \textbf{66.07} & \textbf{69.57}                & +1.22                                                                              \\
    \bottomrule
\end{tabular}
    \caption{Ablation of components in \texttt{AugPT}.}
    \label{tab4}
\end{table}

\subsection{Ablation Study} \label{sec4.3}

Herein, we analyze the effect of each component in \texttt{AugPT}. To enhance the significance of the ablation study, experiments are carried out under full base-to-new setting on 3 datasets, StanfordCars, DTD, and FGVCAircraft, which reveal the most notable improvements. More results are provided in the \textit{Supplementary Material}.

\paragraph{Validity of Proposed Components.}

Tab.~\ref{tab4} presents the performance when gradually integrating \texttt{AugPT}'s \textit{Adaptive Self-supervised Augmentation} (\textbf{ASA}) and \textit{Consensus-based Filtering Gate} (\textbf{CFG}) into the distillation-based backbone model. Since the \textit{Optimized Prompt Distillation} is specifically designed to operate alongside the augmentation module (it would degrade to the original backbone pipeline if isolated), it is not discussed separately. It can be observed that: (1) Introducing the ASA module leads to a significant performance boost, indicating the effectiveness of internal data augmentation strategy proposed by \texttt{AugPT}. This illustrates that further exploiting internal features can effectually reinforce distillation-based prompt tuning. (2) The addition of the CFG module further improves results, suggesting that the Consensus Test successfully filters noisy samples from the augmented views, thereby improving the overall quality of input images.
\paragraph{Influence of Augmentation Strategies.} 
To further validate the effectiveness of the augmentation strategy (\textbf{ASA}) proposed in \texttt{AugPT}, we replace ASA with other mainstream self-supervised augmentation methods for comparison. For fairness, the number of augmented views $N$ and all other sub-modules in \texttt{AugPT} maintain unchanged. As in Tab.~\ref{tab5}, a key conclusion is that, compared to RandAugment \cite{cubuk2020randaugment} with fixed amplitude $A$, \texttt{AugPT} with dynamic amplitude sampled from $A^{(s)} \sim \text{U}(0, A_{\text{max}})$ performs more stable improvements, suggesting better generalization across diverse data distributions. Meanwhile, \texttt{AugPT} outperforms the common Random Resized Crop \cite{mao2022randomresizedcrop} in prompt tuning data processing and achieves performance better than FixMatch \cite{sohn2020fixmatch}.

\begin{table}
\small
\centering
\setlength{\tabcolsep}{2.5pt}

\begin{tabular}{l|ccc|cc} 
\toprule
\rowcolor{gray!10} {\cellcolor{gray!10}}                                            & \multicolumn{3}{c|}{\textbf{HM Performance}}                             & {\cellcolor{gray!10}}                                   & {\cellcolor{gray!10}}                     \\
\rowcolor{gray!10} \multirow{-2}{*}{{\cellcolor{gray!10}}\textbf{Augment Strategy}} & Cars                   & DTD                    & Aircraft               & \multirow{-2}{*}{{\cellcolor{gray!10}}\textbf{Average}} & \multirow{-2}{*}{{\cellcolor{gray!10}}$\Delta$}  \\ 
\midrule
(baseline)                                                                                                        & 83.35                  & 76.38                  & 44.63                  & 68.12                                                                  &                                                          \\
Random Resized Crop                                                                                        & 83.72             & 77.44             & 44.94             & 68.70                                                                     & +0.58                                                     \\
RandAugment                                                                                                       & 83.58             & 78.00             & 44.71             & 68.76                                                                    & +0.64                                                     \\
Fixmatch                                                                                                          & 83.92             & 78.14             & 45.13             & 69.06                                                                    & +0.94                                                     \\
\rowcolor{gray!20}
\textbf{ASA in AugPT} (ours)                                                                                             & \textbf{84.33} & \textbf{78.27} & \textbf{45.69} & \textbf{69.43}                                                         & +1.31                                                    \\
\bottomrule
\end{tabular}
    \caption{Effect of different self-supervised image augmentation strategies on HM performance (\%).}
    \label{tab5}
\end{table}
\begin{figure}
  \centering
  \includegraphics[width=\linewidth]{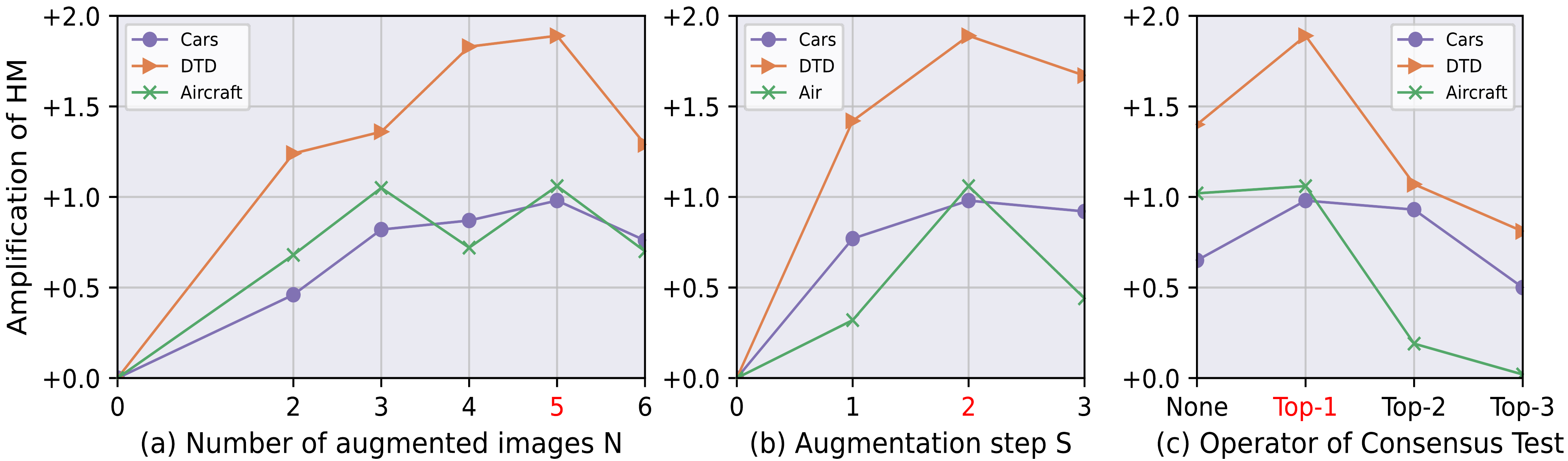}
  \caption{Relative to the backbone model, the HM performance improvement under different (a) augmented image number $N$, (b) step size $S$ and (c) filtering gate operator. }
  \label{Figure 3}
\end{figure}

\paragraph{Size of Augmented Image and Step.}
Fig.~\ref{Figure 3}(a) and (b) evaluate the impact of the number of augmented views $N$ and augmentation step size $S$ in the ASA module on HM performance of \texttt{AugPT}. Across all 3 datasets, performance peaks at $N = 5$ and $S = 2$. For the performance decline when $N > 5$, we believe that this is due to the redundant feature computations caused by excessive augmented views, which may cause overfitting. Similarly, larger step sizes tend to bring in more noise into the raw image, resulting in a decrease of the significance of its subject visual features.

\paragraph{Impact of Stringency in Consensus Test.}
As a comparison to the Top-1 Consensus Test operator used in CFG module of \texttt{AugPT}, we further explore a stricter variant by extending Top-1 to \textit{Top-K} consensus, i.e., only images with consistent Top-K indices of predicted logits would be accepted. As illustrated in Fig.~\ref{Figure 3}(c), increasing the strictness leads to a falloff in performance. We attribute this to the fact that the Top-K Consensus Test discards too many augmented views, compromising the completeness of the input visual features. More variants are in \textit{Supplementary Material B.4}.

\paragraph{Computational Cost.}
Since \texttt{AugPT} avoids the data bottleneck caused by querying LLMs, and introduces no additional learnable parameters, it maintains low computational overhead (e.g., GPU memory usage, tuning Frame Per Second (FPS)) relative to backbones. In terms of inference speed, \texttt{AugPT} is significantly faster than LLM-guided models. Details are in \textit{Supplementary Material B.6}.

\section{Conclusion}
We propose \texttt{AugPT}, an optimized distillation-based prompt tuning method driven by \textbf{internal data augmentation}, requiring no more external data into backbone. It employs \textit{Adaptive Self-supervised Augmentation} to produce diverse views that enrich feature representation of raw images, followed by \textit{Consensus-based Filtering Gate} to discard noisy samples and \textit{Optimized Prompt Distillation} to refine the backbone. Experiments reveal that \texttt{AugPT} surpasses SOTA in multiple tasks, especially in low-data regimes. \textbf{Limitations} and \textbf{future work} are in the \textit{Supplementary Material}.

\section*{Acknowledgements}
This work is supported by the China Scholarship Council (CSC) and the UTS Top-Up Scholarship. Computational facilities are provided by the UTS eResearch High Performance Compute Facilities and the Shanghai Technical Service Computing Center of Science and Engineering, Shanghai University.

\bibliography{aaai2026}


\newpage

\input{sup/Appendix}

\end{document}

%% file: sup/Appendix.tex
\clearpage

\setcounter{secnumdepth}{2}
\renewcommand{\thesection}{\Alph{section}}  
\renewcommand{\thesubsection}{\thesection.\arabic{subsection}}  

\appendix

\begin{strip}
\centering
    {\LARGE \textbf{Raw Data Matters: Enhancing Prompt Tuning by Internal Augmentation on Vision-Language Models}\\[1em]
    \large \textit{Supplementary Material}}\\[2em]
\end{strip}

\texttt{AugPT} is an optimized distillation-based prompt tuning approach that leverages Adaptive Self-supervised Augmentation (ASA), a Consensus-based Filtering Gate (CFG), and Optimized Prompt Distillation (OPD) to extract features from raw images through internal augmentation. As a result, it eliminates the need for additional external data or learnable parameters.

Herein, we provide further details on \texttt{AugPT}. Overall, the Supplementary Material includes:

\begin{itemize}
    \item \textbf{Appendix~\ref{sec:A}}: Additional implementation details, along with descriptions of \texttt{AugPT}’s teacher model and pre-defined augmentation policies in Adaptive Self-supervised Augmentation module.
    \item \textbf{Appendix~\ref{sec:B}}: Extended experiments, including comparisons with additional baselines, error bar analysis, adaptability, computational cost, and further validation of each module proposed in \texttt{AugPT}.
    \item \textbf{Appendix~\ref{sec:C}}: Discussion on societal impact of \texttt{AugPT}.
    \item \textbf{Appendix~\ref{sec:D}}: Limitations of \texttt{AugPT} and our planned future work.
\end{itemize}

\section{More Implementation Details} \label{sec:A}
Herein, we provide additional detailed setup of \texttt{AugPT} to enhance the reproducibility of our model.

\subsection{Experimental Setup} \label{sec:A1}

\paragraph{Datasets.}
Following the standard paradigm of mainstream prompt tuning approaches, \texttt{AugPT} is fine-tuned and evaluated on 11 datasets across the 3 downstream tasks described in \textit{Experiments} section. Details of the datasets, along with the unlabeled images in training set introduced for each fine-tuning task, are summarized in Tab.~\ref{tabS1}.

As a distillation-based approach, the data sampling strategy for the fine-tuning and inference process of \texttt{AugPT} remains consistent with the backbone model PromptKD \cite{li2024promptkd}.

\begin{table}[h]
\centering
\small
\setlength\tabcolsep{4pt}

\begin{tabular}{c|ccc|cc} 
\toprule
\rowcolor{gray!10} {\cellcolor{gray!10}}                                   & \multicolumn{3}{c|}{\textbf{Training Set}} & \multicolumn{2}{c}{\textbf{Test Set}}  \\  
\rowcolor{gray!10} \multirow{-2}{*}{{\cellcolor{gray!10}}\textbf{Dataset}} & 8-shot & 16-shot & Full            & Base  & New                        \\  
\midrule
ImageNet                                                                                                 & 8000   & 16000   & 1281167         & 25000 & 25000                                                                \\
Caltech101                                                                                               & 800    & 1600    & 4128            & 1549  & 916                                                                   \\
OxfordPets                                                                                               & 296    & 592     & 2944            & 1881  & 1788                       \\
StandfordCars                                                                                            & 1568   & 3136    & 6509            & 4002  & 4039                       \\
Flowers102                                                                                               & 816    & 1632    & 4093            & 1053  & 1410                       \\
Food101                                                                                                  & 808    & 1616    & 50500           & 15300 & 15000                      \\
FGVCAircraft                                                                                             & 800    & 1600    & 3334            & 1666  & 1667                       \\
SUN397                                                                                                   & 3176   & 6352    & 15880           & 9950  & 9900                       \\
DTD                                                                                                      & 376    & 752     & 2820            & 864   & 828                        \\
EuroSAT                                                                                                  & 80     & 160     & 13500           & 4200  & 3900                       \\
UCF101                                                                                                   & 808    & 1616    & 7639            & 1934  & 1849                       \\
\bottomrule
\end{tabular}
    \caption{Detailed information of 11 datasets utilized in prompt tuning. }
    \label{tabS1}
\end{table}

\paragraph{Hyperparameters.}
Following the setup of the PromptKD backbone, as the initialization setup, we set the the length of learnable text or image prompts in the teacher and student model to 4, as well as the depth to 9 (insert prompts between the first 9 Transformer blocks in ViT encoders). The learnable image prompts of teacher and student model are randomly initialized adhering to a zero-mean Gaussian distribution $X \sim \mathcal{N}(0,\ 0.02^2)$. The learnable feature projection layer $\text{Proj}(\cdot)$ in student model is initialized randomly with 2 MLP layers (ablation study of layers is in Sec.~\ref{sec:B4}). The fine-tuning details of the applied teacher model are introduced in Sec.~\ref{sec:A2}. 

During the fine-tuning stage of the student model, we utilize stochastic gradient descent (SGD) optimizer, fine-tuning for 20 epochs with batch size $bs=16$ and learning rate $lr=0.005$. We use the default Random Resized Crop and Random Flip as the initial image transformation strategies for raw images. The temperature hyperparameter in the distillation-based method is kept as the default value of $\tau =1$. 

All experiments are conducted on a single NVIDIA A40 GPU in the internal cluster with 128 GB memory and 1 TB allocated storage.

\subsection{More Details of Distillation-based Prompt Tuning Backbone} \label{sec:A2}
\texttt{AugPT} introduces PromptSRC \cite{khattak2023promptsrc} pre-tuned by PromptKD using ViT-L/14 encoders as a teacher model. As a supplement to \textit{Preliminaries} section (Sec. 3.1) in main text, we provide more detailed description of the teacher model and the fine-tuning process associated with the standard paradigm of Distillation-based Prompt Tuning approaches to which \texttt{AugPT} belongs.

\begin{figure*}[t]
  \centering
  \includegraphics[width=\textwidth]{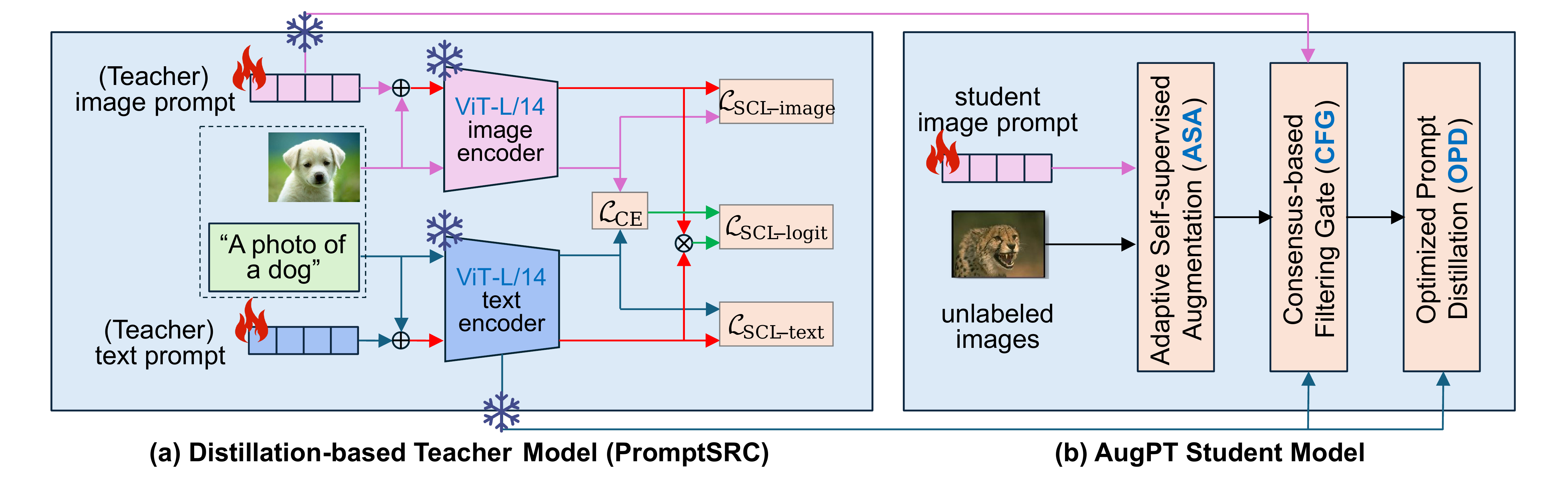}
  \caption{The framework of (a) the teacher model used in distillation-based prompt tuning approaches. In (b) \texttt{AugPT}, the pre-tuned teacher image prompts $\boldsymbol{P}_{v}^{\text{tea}}$ and text prompts $\boldsymbol{P}_{t}^{\text{tea}}$ are utilized for online inference in Consensus-based Filtering Gate (Sec. 3.3 in main text). Moreover, textual features obtained by ViT-L/14-based teacher text modality are used in Optimized Prompt Distillation (Sec. 3.4 in main text) for interacting with student visual features. }
  \label{Figure 4}
\end{figure*}

\paragraph{Teacher Model.}
PromptSRC adopts the IVLP \cite{rasheed2023ivlp} (Independent Visual-Language Prompting) framework by separately optimizing visual and textual prompts during fine-tuning. To alleviate the performance degradation caused by the Base-to-New Trade-off (BNT) problem \cite{zhou2022cocoop, zhang2024dept, li2025dpc}, it introduces a series of structurally aligned regularization losses. Beyond the conventional cross-entropy loss $\mathcal{L}_{\text{CE}}$ commonly imported in mainstream prompt tuning methods, PromptSRC enforces additional consistency constraints across modalities, which are defined as $\mathcal{L}_{\text{SCL-image}}$ and $\mathcal{L}_{\text{SCL-text}}$, encouraging better alignment between prompts and their respective modality-specific features, while $\mathcal{L}_{\text{SCL-logits}}$ further regularizes the logits after cross-modal interaction. These jointly contribute to improved generalization across both base and new tasks. The framework of PromptSRC is illustrated in Fig.~\ref{Figure 4}. The optimizing goal of fine-tuning can be summarized as follows: by continuously optimizing the weighted sum of the cross-entropy loss $\mathcal{L}_{\text{CE}}$ and the consistency loss  $\mathcal{L}_{\text{SCL}}$ , the output logits obtained from the learnable prompts as input queries are progressively aligned with the one-hot distribution of the ground-truth label.

\begin{equation}
    \begin{aligned}
        \mathcal{L} 
        & \, =\mathcal{L}_{\text{CE}} + \mathcal{L}_{\text{SCL}} \\
        & \, = \mathcal{L}_{\text{CE}} +\lambda_{1}\mathcal{L}_{\text{SCL-image}} + \lambda_{2}\mathcal{L}_{\text{SCL-text}} + \mathcal{L}_{\text{SCL-logits}}
    \end{aligned}
\end{equation}
where $\lambda_{1}$ and $\lambda_{2}$ are hyperparameters used as weight coefficients.

\paragraph{Settings of Distillation-based Prompt Tuning.}
In the fine-tuning stage, serving as the teacher model in distillation-based prompt tuning approaches, PromptSRC \cite{khattak2023promptsrc} adopts a pre-trained CLIP model with larger ViT-L/14 encoders as the foundation model. At initialization, PromptSRC uses the prompt template ``A photo of a [CLS]'' to initialize the learnable text prompts and samples the learnable image prompts from a zero-mean Gaussian distribution $X \sim \mathcal{N}(0,\ 0.02^2)$. Both text and image prompts have a depth of 9 and a shape of $[4, 768]$.

For the fine-tuning process on base-to-new tasks, the PromptSRC teacher model is tuned using few-shot image-text pairs sampled from base classes. The pre-defined hyperparameters are $epoch = 20$, $bs = 16$, and $lr = 0.002$, using the same SGD optimizer. Additionally, the loss weights are set as $\lambda_1 = 10$ and $\lambda_2 = 25$.

When fine-tuned PromptSRC is applied as teacher model for inference, the matching probability between image features $f^{\text{tea}}(\boldsymbol{P}_v, I)$ and text features $g^{\text{tea}}(\boldsymbol{P}_t)$ from candidate set $\boldsymbol{C} = \{ T_i \}_{i=1}^{c}$ is:

\begin{equation}
    p(y \mid I)=\frac{\exp \left(sim[g^{\text{tea}}({\boldsymbol{P}_t}_y), f^{\text{tea}}(\boldsymbol{P}_v,I)] / \tau\right)}{\sum_{i=1}^{c} \exp \left(sim[g^{\text{tea}}(\boldsymbol{P}_{t_i}), f^{\text{tea}}(\boldsymbol{P}_v,I)]/ \tau\right)}
\end{equation}

where $\text{sim}[\cdot, \cdot]$ denotes cosine similarity and $\tau$ is the temperature coefficient. It is worth noting that, following the PromptKD \cite{li2024promptkd} setting, the candidate set $\boldsymbol{C}$ includes all class names, containing both base and new classes. Since only unlabeled images are introduced during fine-tuning as raw data, the relationship between class labels and images is entirely unknown in the fine-tuning process, ensuring that no data leakage occurs.

The fine-tuning method for the student model is described in \textit{Preliminaries} section (Sec. 3.1). The shape of 9 student image prompts are $[4, 512]$. \texttt{AugPT} further introduces an optimized variant of this procedure in \textit{Method} section (Sec. 3.4) in main text, referred to as Optimized Prompt Distillation (OPD).

\subsection{More Details of Adaptive Self-supervised Augmentation} \label{sec:A3}
\texttt{AugPT} follows the setup of RandAugment \cite{cubuk2020randaugment} by independently sampling $S$ augmentation operations from a pre-defined set of mainstream image transformation policies $\mathcal{Y} = \{\gamma_1, \dots, \gamma_Q\}$ for constructing a dynamic policy group $\mathcal{G}$ on a single augmentation, then performs $N$ self-supervised augmentation on the raw unlabeled image $I'$ to obtain the augmented set $\mathcal{D}(I')$ with $N+1$ views. The complete list of $Q = 16$ candidate augmentation operations along with their corresponding magnitude ranges $(A_{\text{min}}, A_{\text{max}})$ is provided in Tab.~\ref{tabS2}.

\begin{table*}[t]
\centering
\small
\setlength\tabcolsep{3pt}

\begin{tabular}{l|cccc} 
\toprule
{\cellcolor{gray!10}}\textbf{Transformation Policy $\mathcal{Y}$} & \makebox[0.15\textwidth][c]{AutoContrast} & \makebox[0.15\textwidth][c]{Equalize}   & \makebox[0.15\textwidth][c]{Invert}     & \makebox[0.15\textwidth][c]{Rotate}       \\ 
\midrule
{\cellcolor{gray!10}}\textbf{Amplitude interval $(A_)$}          & (0, 1)       & (0, 1)     & (0, 1)     & (0, 30)      \\ 
\midrule
\midrule
{\cellcolor{gray!10}}\textbf{Transformation Policy $\mathcal{Y}$} & Posterize    & Cutout     & Solarize   & SolarizeAdd  \\ 
\midrule
{\cellcolor{gray!10}}\textbf{Amplitude $A$}          & (4, 8)       & (0, 0.2)   & (0, 256)   & (0, 110)     \\ 
\midrule
\midrule
{\cellcolor{gray!10}}\textbf{Transformation Policy $\mathcal{Y}$} & Color        & Contrast   & Brightness & Sharpness    \\ 
\midrule
{\cellcolor{gray!10}}\textbf{Amplitude $A$}          & (0.1, 1.9)   & (0.1, 1.9) & (0.1, 1.9) & (0.1, 1.9)   \\ 
\midrule
\midrule
{\cellcolor{gray!10}}\textbf{Transformation Policy $\mathcal{Y}$}  & ShearX       & ShearY     & TranslateX & TranslateY   \\ 
\midrule
{\cellcolor{gray!10}}\textbf{amplitude $A$}          & (0, 0.3)     & (0, 0.3)   & (0, 0.33)  & (0, 0.33)    \\
\bottomrule
\end{tabular}
    \caption{Detailed information of pre-defined image transformation policies in RandAugment. }
    \label{tabS2}
\end{table*}

\begin{figure*}[t]
  \centering
  \includegraphics[width=0.9\textwidth]{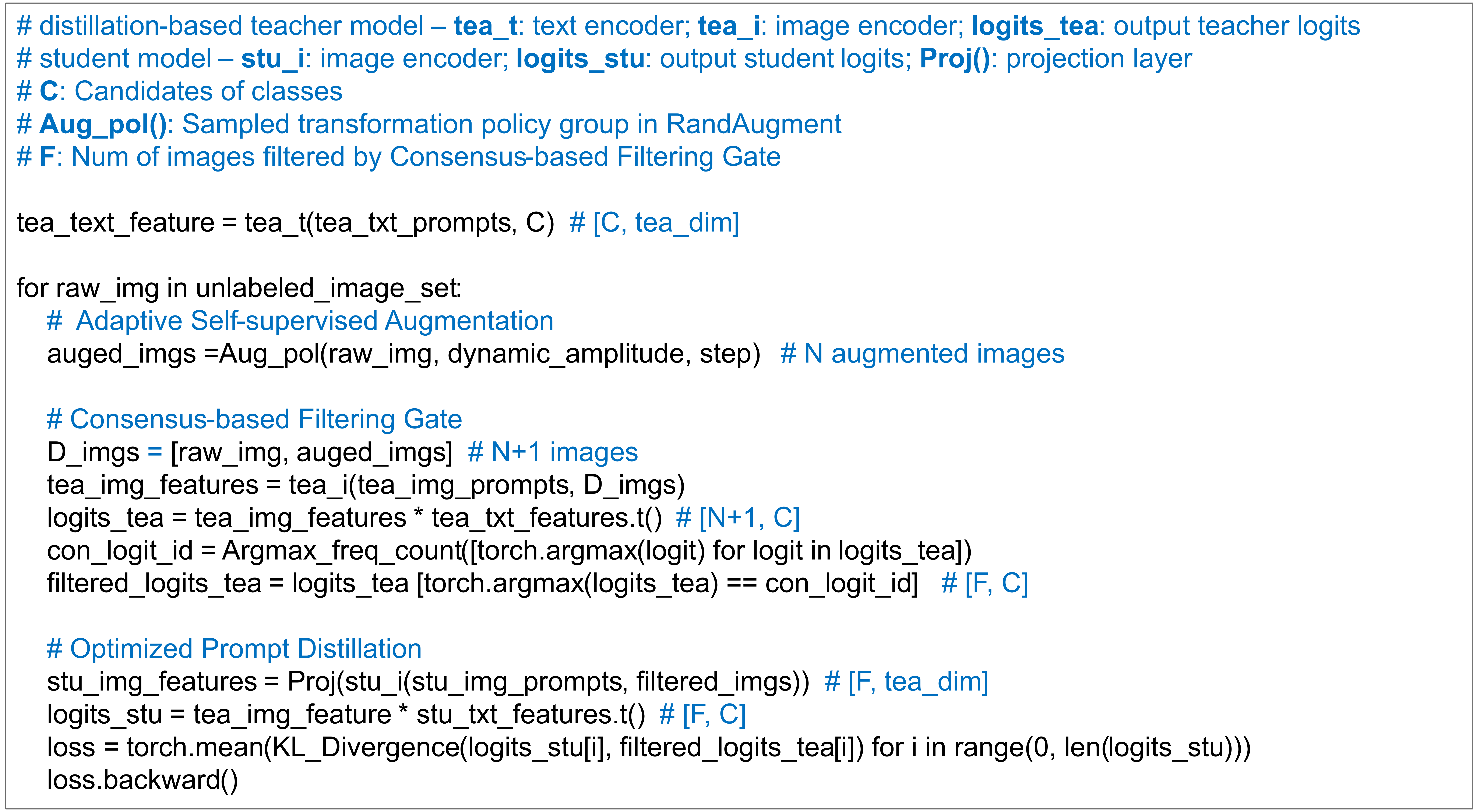}
  \caption{Pseudo-code of \texttt{AugPT} in PyTorch.}
  \label{Figure 5}
\end{figure*}

For the fixed amplitude hyperparameter $A$ pre-searched by RandAugment (on CIFAR10 \cite{krizhevsky2009cifar10}), the transformation strength for each image operation $\hat{A}$ is determined by the following conversion:

\begin{equation} \label{eqn14}
\hat{A}= \frac{A}{30}\left(A_{\text{max}}-A_{\text{min}}\right)+A_{\text{min}}
\end{equation}

In contrast, \texttt{AugPT} introduces a uniformly sampled magnitude $A^{(s)} \sim \text{U}(0, A_{\text{max}})$ to improve the generalization ability of self-supervised augmentation across diverse data distributions, removing the demand for heuristic search in RandAugment, signifying less computational overhead. Accordingly, in \texttt{AugPT}, the formulation of Eqn.~\ref{eqn14} is modified to:

\begin{equation} \label{eqn15}
\hat{A}= \frac{A^{(s)}}{A_{\text{max}}} \left(A_{\text{max}}-A_{\text{min}}\right)+A_{\text{min}}
\end{equation}

\subsection{Pseudo-code of Algorithm} \label{sec:A4}
Here, we provide PyTorch-style pseudo-code for \texttt{AugPT} in Fig.~\ref{Figure 5} as a demonstration. We promise to release the complete implementation in the future, including pre-tuned weights of the teacher model on 11 datasets, as well as the full code for fine-tuning and evaluating \texttt{AugPT}.

\begin{table*}[t]
    \centering
    \small
    \setlength\tabcolsep{5.5pt}
\begin{tabular}{cl|ccc} 
\toprule
\rowcolor{gray!10} {\cellcolor{gray!10}}                                                                                                                                  & {\cellcolor{gray!10}}                                 & \multicolumn{3}{c}{\textbf{Avg. over 11 datasets}}                                                                     \\
\rowcolor{gray!10} \multirow{-2}{*}{{\cellcolor{gray!10}}\begin{tabular}[c]{@{}>{\cellcolor{gray!10}}c@{}}\textbf{Tuning}\\\textbf{Strategy}\end{tabular}} & \multirow{-2}{*}{{\cellcolor{gray!10}}\textbf{Model}} & Base                                  & New                                   & H                                      \\ 
\midrule
zero-shot                                                                                                                                                                                               & CLIP \cite{radford2021clip}                                                                 & 69.34                                 & 74.22                                 & 71.70                                  \\ 
\midrule
\multirow{5}{*}{few-shot}                                                                                                                                                                               & CoCoOp \cite{zhou2022cocoop}                                                               & 80.47                                 & 71.69                                 & 75.83                                  \\
                                                                                                                                                                                                        & KgCoOp \cite{yao2023kgcoop}                                                               & 80.73                                 & 73.60                                 & 77.00                                  \\
                                                                                                                                                                                                        & MaPLe \cite{khattak2023maple}                                                               & 82.28                                 & 75.14                                 & 78.55                                  \\
                                                                                                                                                                                                        & TCP \cite{yao2024tcp}                                                                 & 84.13                                 & 75.36                                 & 79.51                                  \\
                                                                                                                                                                                                        & MMRL \cite{guo2025mmrl}                                                                & 85.68                                 & 77.16                                 & 81.20                                  \\ 
\midrule
\multirow{2}{*}{\begin{tabular}[c]{@{}c@{}}few-shot +\\unlabeled images\end{tabular}}                                                                                                                   & MAO \cite{li2025mao}                                                                 & 84.53                                 & 75.38                                 & 79.69                                  \\
                                                                                                                                                                                                        & CasPL \cite{wu2024caspl}                                                               & 86.11                                 & 79.54                                 & 82.69                                  \\ 
\midrule
\multirow{3}{*}{unlabeled images}                                                                                                                                                                       & KDPL \cite{mistretta2024kdpl}                                                                 & 77.11                                 & 71.61                                 & 74.26                                  \\
& \textbf{AugPT} (16-shot)  & 83.95 & 78.59 & 81.18 \\
                                                                                                                                                                                                        & {\cellcolor{cyan!15}}\textbf{AugPT} (Full)                                & {\cellcolor{cyan!15}}\textbf{87.31} & {\cellcolor{cyan!15}}\textbf{80.87} & {\cellcolor{cyan!15}}\textbf{83.97}  \\
\bottomrule
\end{tabular}
    \caption{Comparison of \texttt{AugPT} with more baselines on base-to-new tasks over 11 datasets.}
    \label{tabS3}
\end{table*}

\begin{table}[h]
    \centering
    \small
\setlength\tabcolsep{5pt}
\begin{tabular}{c|ccc}
\toprule
\rowcolor{gray!10} \textbf{Method} & \textbf{Average} & \textbf{ImageNet} & \textbf{Caltech101} \\
\midrule
PromptKD & 83.41 & 77.62 & 97.81 \\
\rowcolor{cyan!15}
\textbf{AugPT} & 83.97 (±0.09) & 77.84 (±0.04) & 97.96 (±0.05) \\
\midrule
\rowcolor{gray!10} \textbf{Method} & \textbf{OxfordPets} & \textbf{StanfordCars} & \textbf{Flowers102} \\
\midrule
PromptKD & 97.18 & 83.35 & 89.72 \\
\rowcolor{cyan!15}
\textbf{AugPT} & 97.53 (±0.05) & 84.33 (±0.01) & 90.21 (±0.02) \\
\midrule
\rowcolor{gray!10} \textbf{Method} & \textbf{Food101} & \textbf{FGVCAircraft} & \textbf{SUN397} \\
\midrule
PromptKD & 93.25 & 44.63 & 82.36 \\
\rowcolor{cyan!15}
\textbf{AugPT} & 93.45 (±0.06) & 45.69 (±0.17) & 82.75 (±0.09) \\
\midrule
\rowcolor{gray!10} \textbf{Method} & \textbf{DTD} & \textbf{EuroSAT} & \textbf{UCF101} \\
\midrule
PromptKD & 76.38 & 88.23 & 85.09 \\
\rowcolor{cyan!15}
\textbf{AugPT} & 78.27 (±0.15) & 88.77 (±0.20) & 85.37 (±0.25) \\
\bottomrule
\end{tabular}
\caption{Error bar analysis of \texttt{AugPT} on base-to-new tasks over 11 datasets, with average scores shown in the top row.}
\label{tabS4}
\end{table}

\section{More Experimental Results} \label{sec:B}

\subsection{Compare with More Baselines} \label{sec:B1}

To further compare the performance of \texttt{AugPT} against a wider range of prompt tuning models, we introduce additional baseline models in Tab.~\ref{tabS3}. These models include the original zero-shot CLIP \cite{radford2021clip}, as well as CoCoOp \cite{zhou2022cocoop}, KgCoOp \cite{yao2023kgcoop}, MaPLe \cite{khattak2023maple}, TCP  \cite{yao2024tcp}, and MMRL \cite{guo2025mmrl}, which are all fine-tuned with few-shot image-text pairs. Additionally, we incorporate MAO \cite{li2025mao} and CasPL \cite{wu2024caspl}, which utilize a combination of few-shot image-text pairs and unlabeled images. Another distillation-based prompt tuning approach KDPL \cite{mistretta2024kdpl} is also contained in the comparison.

It can be observed that \texttt{AugPT} requires only 16-shot unlabeled images to achieve performance comparable to advanced prompt tuning models \cite{guo2025mmrl} fine-tuned with the same shot of image-text pairs, indicating a lower data requirement. Similarly, when fine-tuned on full dataset, \texttt{AugPT} outperforms models \cite{wu2024caspl} that use both few-shot image-text pairs and full unlabeled images.

\begin{table}[t]
    \centering
    \small
    \setlength\tabcolsep{4pt}
    
\begin{tabular}{c|ccc|ccc}
\toprule
\rowcolor{gray!10} {\cellcolor{gray!10}}                                   & \multicolumn{3}{c|}{\textbf{N=0 (baseline)}} & \multicolumn{3}{c}{\textbf{N=2}}  \\
\rowcolor{gray!10} \multirow{-2}{*}{{\cellcolor{gray!10}}\textbf{Dataset}}
 & Base & New & H & Base & New & H \\
\midrule
StanfordCars   & 82.61 & 84.10 & 83.35 & 83.61 & 84.01 & 83.81 \\
DTD            & 86.34 & 68.48 & 76.38 & 86.11 & 70.65 & 77.62 \\
FGVCAircraft   & 49.10 & 40.91 & 44.63 & 49.34 & 41.89 & 45.31 \\
\midrule
\rowcolor{gray!10} {\cellcolor{gray!10}}                                   & \multicolumn{3}{c|}{\textbf{N=3}} & \multicolumn{3}{c}{\textbf{N=4}}  \\
\rowcolor{gray!10} \multirow{-2}{*}{{\cellcolor{gray!10}}\textbf{Dataset}}
& Base & New & H & Base & New & H \\
\midrule
StanfordCars   & 83.83 & 84.50 & 84.16 & 83.88 & 84.55 & 84.21 \\
DTD            & 86.23 & 70.77 & 77.74 & 87.04 & 71.01 & 78.21 \\
FGVCAircraft   & 49.10 & 42.71 & 45.68 & 49.22 & 42.05 & 45.35 \\
\midrule
\rowcolor{gray!10} {\cellcolor{gray!10}}                                   & \multicolumn{3}{c|}{\textbf{N=5}} & \multicolumn{3}{c}{\textbf{N=6}}  \\
\rowcolor{gray!10} \multirow{-2}{*}{{\cellcolor{gray!10}}\textbf{Dataset}}
& Base & New & H & Base & New & H \\
\midrule
StanfordCars   & 84.31 & 84.35 & \textbf{84.33} & 83.91 & 84.30 & 84.10 \\
DTD            & 86.81 & 71.26 & \textbf{78.27} & 86.23 & 70.65 & 77.67 \\
FGVCAircraft   & 49.28 & 42.59 & \textbf{45.69} & 49.10 & 42.11 & 45.34 \\
\bottomrule
\end{tabular}
    \caption{Detailed results of ablation under different augmented images $N$ on full base-to-new tasks.}
    \label{tabS5}
\end{table}

\subsection{Error Bar Analysis} \label{sec:B2}
Given that \texttt{AugPT}'s internal data augmentation introduces a degree of randomness into the process of prompt tuning, we conduct an error bar analysis on the main base-to-new task across all 11 datasets. We fix all hyperparameters and the random seed in the model, and perform 3 runs per dataset. Tab.~\ref{tabS4} reports both the mean and the standard deviation of Top-1 accuracy.

\begin{table}[t]
    \centering
    \small
\setlength\tabcolsep{2.5pt}

\begin{tabular}{cl|ccc|c} 
\toprule
\rowcolor{gray!10} {\cellcolor{gray!10}}                   & {\cellcolor{gray!10}}                                            & \multicolumn{3}{c|}{\textbf{HM Performance}}  & {\cellcolor{gray!10}}                                   \\
\rowcolor{gray!10} \multirow{-2}{*}{{\cellcolor{gray!10}}} & \multirow{-2}{*}{{\cellcolor{gray!10}}\textbf{Image Processor}} & Cars & DTD & Aircraft & \multirow{-2}{*}{{\cellcolor{gray!10}}\textbf{Average}} \\ 
\midrule
                                                                                         & (baseline)                 & 83.35 & 76.38 & 44.63 & 68.12 \\
(1)                                                                                      & Copy $N$ images            & 83.34 & 76.64 & 44.71 & 68.23 \\
(2)                                                                                      & $N$ Random Resized Crop    & 83.72 & 77.44 & 44.94 & 68.70 \\
\rowcolor{cyan!15}
(3)                                                                                      & \textbf{AugPT} (full ASA module) & \textbf{84.33} & \textbf{78.27} & \textbf{45.69} & \textbf{69.43} \\
\bottomrule
\end{tabular}
    \caption{Detailed results of ablation under different \textbf{image processors} on full base-to-new tasks.}
    \label{tabS6}
\end{table}

\subsection{Effectiveness of Adaptive Self-supervised Augmentation} \label{sec:B3}

\paragraph{Detailed Results on Augmented Views $N$.}
To further evaluate the performance trend of the ASA module (\textit{Method} section in main text) in \texttt{AugPT} under different numbers of augmented images $N$, we provide the detailed numerical results corresponding to Fig.~\ref{Figure 5}(a) in Tab.~\ref{tabS5}.

\paragraph{Further Verification of the Effectiveness of ASA.}
To further verify that the performance improvement of \texttt{AugPT} does not stem from simple data stacking, we introduce an additional experiment to evaluate the effectiveness of the image augmentation strategy proposed by ASA module. Specifically, under the original setting of $N = 5$ in \texttt{AugPT}, we compare 3 strategies: (1) replacing all $N$ augmented images with repeated copies of the raw image, (2) applying $N$ rounds of Random Resized Crop, which is the common image transformation method in prompt tuning backbones (\textit{Ablation Study} section (Sec. 4.3) in main text), and (3) applying ASA-based augmentation for $N$ times in \texttt{AugPT}.

As shown in Tab.~\ref{tabS6}, simply duplicating the image results in only marginal or even degraded performance, likely due to overfitting caused by repeated feature inputs during fine-tuning. Similarly, \texttt{AugPT} outperforms the strategy of applying basic transformations alone. These results confirm that the augmented images output by the ASA module can more effectively capture the latent features of the raw images through more diverse views.

\subsection{Effectiveness of Consensus-based Filtering Gate} \label{sec:B4}

\paragraph{Theoretical Explanation of Consensus Test.}
In the \textit{Consensus-based Filtering Gate} section (Sec. 3.3) of main text, we outline the mechanism of the Consensus Test based on pre-tuned teacher model, where the consistency of top-1 logit predictions can emerge as a reliable signal for evaluating visual similarity because that the augmented views are encoded by the uniform frozen prompt tuning backbone. Herein, we provide further theoretical analysis to enhance the interpretability of this mechanism.

The phenomenon that images exhibiting consistent top-1 logit predictions tend to share visual similarity can be theoretically interpreted via the embedding geometry induced by the ViT-based encoders of CLIP. Specifically, both the encoders ($f^{\text{tea}}(\cdot)$ and $g^{\text{tea}}(\cdot)$) and the tuned prompt vectors ($\boldsymbol{P}_{v}^{\text{tea}}$ and $\boldsymbol{P}_{t}^{\text{tea}}$), are frozen in the teacher model. Therefore, for the feature extraction process on an unlabeled image $I'$ :

\begin{equation}
\boldsymbol{e}_{I'} = f^{\text{tea}}({\boldsymbol{P}_{v}^{\text{tea}}},I'),\quad I' \in \mathcal{D}(I')
\end{equation}

it can be abstracted as a \textit{deterministic mapping with fixed parameters}.

\begin{equation}
h_{\alpha}: I' \rightarrow \boldsymbol{e}_{I'}
\end{equation}

where $h_{\alpha}$ denotes the fixed parameter set of the teacher model. Combining Eqn.~\ref{eqn6} and Eqn.~\ref{eqn7} in the main text, for a single unlabeled image $I'$, the top-1 prediction based on the candidate text features $\boldsymbol{e}_{c} = g^{\text{tea}}(\boldsymbol{P}_{t}^{\text{tea}})$ can be formulated as:

\begin{equation}
c^* = \underset{c \in \boldsymbol{C}}{\mathrm{argmax}}\ sim[h_{\alpha}(I'),\boldsymbol{e}_{c}]
\end{equation}

When multiple augmented images $\{I'_j\}_{j=1}^{\hat{N}\leq N+1}$ yield identical top-1 predictions (in AugPT, $\hat{N}=|\mathcal{I}_{acc}|$ as mentioned in \textit{Optimized Prompt Distillation} section (Sec. 3.4) in the main text), we have:

\begin{equation}
\forall j,\quad \underset{c \in \boldsymbol{C}}{\mathrm{argmax}}\ sim[h_{\alpha}(I'_j),\boldsymbol{e}_{c}] = c^*.
\end{equation}

This implies that augmented embeddings $h_{\alpha}(I'_j)$ reside within the same decision boundary region around the embedding $\boldsymbol{e}_{c^*}$. Hence, their pairwise embedding distances (quantified by cosine similarity) are significantly smaller compared to embeddings of noisy-sample images. Mathematically, this embedding proximity can be stated as:

\begin{equation}
sim[h_{\alpha}(I'_j), h_{\alpha}(I'_k)] \approx 1,\quad \forall j,k \in \{1,\dots,\hat{N}\}
\end{equation}

Since $h_{\alpha}$ is a deterministic mapping with fixed parameters, the consistency in representation-level similarity among ${I'_j}$ implies that these augmented inputs share highly similar semantic content, i.e.:

\begin{equation}
sim[I'_j, I'_k] \approx 1,\quad \forall j,k \in \{1,\dots,\hat{N}\}
\end{equation}

The above process explains the consistency principle between visual feature similarity and image similarity in the prompt tuning. Based on this, \texttt{AugPT} introduces the Consensus-based Filtering Gate, which leverages the teacher model to effectively reserve highly similar image samples as reliable augmented views.

\begin{table}[t]
    \centering
    \small
\setlength\tabcolsep{2pt}

\begin{tabular}{l|ccc|c} 
\toprule
\rowcolor{gray!10} {\cellcolor{gray!10}}                                          & \multicolumn{3}{c|}{\textbf{HM Performance}}     & {\cellcolor{gray!10}}                                    \\
\rowcolor{gray!10} \multirow{-2}{*}{{\cellcolor{gray!10}}\textbf{Filtering Gate}} & Cars           & DTD            & Aircraft       & \multirow{-2}{*}{{\cellcolor{gray!10}}\textbf{Average}}  \\

\midrule
(baseline)                                                                                      & 83.35           & 76.38          & 44.63          & 68.12 \\
zero-shot ViT-B/16 CLIP                                                                         & 83.63           & 78.13          & 45.17          & 68.97 \\
zero-shot ViT-L/14 CLIP                                                                         & 83.93           & 77.72          & 45.01          & 68.89 \\
\rowcolor{cyan!15} \textbf{AugPT} (tuned ViT-L/14 teacher)                                      & \textbf{84.33}  & \textbf{78.27} & \textbf{45.69} & \textbf{69.43} \\
\bottomrule
\end{tabular}
    \caption{Detailed results of ablation under different \textbf{filtering gates} on full base-to-new tasks.}
    \label{tabS7}
\end{table}

\paragraph{Variants of Filtering Gate.}
In In \textit{Ablation Study} section (Sec. 4.3) of main text, we extend the Consensus Test operator in the Filtering Gate from Top-1 to a stricter Top-K form. In this section, we further investigate the effect of a weaker form of the Consensus Test on the performance of \texttt{AugPT}.

Specifically, we replace the pre-tuned teacher model used for online inference in the Consensus Test (i.e., PromptSRC \cite{khattak2023promptsrc} with ViT-L/14 encoders) with the zero-shot foundation CLIP model without fine-tuning. Since the foundation CLIP has no knowledge of the in-domain data distribution, its ability to infer consensus over image features in base tasks is expected to degrade. We experiment with both ViT-L/14 and ViT-B/16 versions of zero-shot CLIP as the Filtering Gate, and execute Top-1 Consensus Test for fair comparison.

Results in Tab.~\ref{tabS7} show that though \texttt{AugPT} with zero-shot CLIP as Filtering Gate still achieves noticeable gains over the baseline, its performance is lower compared to the model introducing pre-tuned teacher model. We believe that this is because the teacher model learns the alignment of images and texts on the target data distribution, which makes it more effective in filtering the augmented views. This experiment further validates the importance of \texttt{AugPT}'s distillation-based prompt tuning design.

\begin{table}[t]
    \centering
    \small
    \setlength\tabcolsep{4pt}
\begin{tabular}{cc|ccc|c} 
\toprule
\rowcolor{gray!10} {\cellcolor{gray!10}}                                     & {\cellcolor{gray!10}}                                  & \multicolumn{3}{c|}{\textbf{Average Performance}}                                                                                                            & {\cellcolor{gray!10}}                                \\
\rowcolor{gray!10} \multirow{-2}{*}{{\cellcolor{gray!10}}\textbf{MLP Layer}} & \multirow{-2}{*}{{\cellcolor{gray!10}}\textbf{Method}} & Base                                               & New                                                & H                                                  & \multirow{-2}{*}{{\cellcolor{gray!10}}\textbf{$\Delta$}}  \\ 
\midrule
\multirow{2}{*}{1}                                                                                         & PromptKD                                                              & 72.40                                              & 64.07                                              & 67.98                                              &                                                                     \\
                                                                                                           & {\cellcolor{cyan!15}}\textbf{AugPT}                    & {\cellcolor{cyan!15}}73.14          & {\cellcolor{cyan!15}}65.61          & {\cellcolor{cyan!15}}69.17          & {\cellcolor{cyan!15}}+1.19                           \\ 
\midrule
\multirow{2}{*}{2}                                                                                         & PromptKD                                                              & 72.68                                              & 64.50                                              & 68.35                                              &                                                                     \\
                                                                                                           & {\cellcolor{cyan!15}}\textbf{AugPT}                    & {\cellcolor{cyan!15}}\textbf{73.47} & {\cellcolor{cyan!15}}\textbf{66.07} & {\cellcolor{cyan!15}}\textbf{69.57} & {\cellcolor{cyan!15}}+1.22                           \\ 
\midrule
\multirow{2}{*}{3}                                                                                         & PromptKD                                                              & 72.52                                              & 63.91                                              & 67.94                                              &                                                                     \\
                                                                                                           & {\cellcolor{cyan!15}}\textbf{AugPT}                    & {\cellcolor{cyan!15}}73.35          & {\cellcolor{cyan!15}}65.60          & {\cellcolor{cyan!15}}69.26          & {\cellcolor{cyan!15}}+1.32                           \\
\bottomrule
\end{tabular}
    \caption{Detailed results of ablation under different MLP layer in feature projector in Optimized Prompt Distillation module across 3 datasets on full base-to-new tasks. }
    \label{tabS8}
\end{table}

\begin{table}[t]
    \centering
    \small
    \setlength\tabcolsep{8pt}
    
\begin{tabular}{c|ccc} 
\toprule
\rowcolor{gray!10} {\cellcolor{gray!10}}                          & \multicolumn{3}{c}{\textbf{Accuracy}}    \\
\rowcolor{gray!10} \multirow{-2}{*}{{\cellcolor{gray!10}}\textbf{Dataset}} & CLIP  & UPL   & \textbf{UPL+AugPT}          \\ 
\midrule
StanfordCars                                                                                    & 55.64 & 70.97 & \textbf{72.03}  \\
DTD                                                                                             & 41.61 & 55.08 & \textbf{56.91}  \\
FGVCAircraft                                                                                    & 16.92 & 21.75 & \textbf{22.11}  \\ 
\midrule
\rowcolor{cyan!15}
\textbf{Average}                                                                                         & 38.06 & 49.27 & \textbf{50.35}  \\
\bottomrule
\end{tabular}

    \caption{Fine-tuning performance of the UPL model using \texttt{AugPT} strategy on 3 typical datasets. }
    \label{tabS10}

\end{table}

\paragraph{Ablation of Projection Layer.}
In Optimized Prompt Distillation, \texttt{AugPT} follows the PromptKD \cite{li2024promptkd} backbone by introducing an MLP-based projection layer $\text{Proj}(\cdot)$ to align student image features with teacher text features. To further verify the potential impact of the number of layers in this projection module, we conduct an ablation study in Tab.~\ref{tabS8} by varying the default 2-layer MLP to 1 or 3 layers.

We observe that, consistent with the PromptKD backbone, \texttt{AugPT} achieves the highest HM performance with a 2-layer projection. Using fewer or more layers may lead to underfitting or overfitting in the alignment process, resulting in degraded performance.

\subsection{Adaptability of AugPT} \label{sec:B6}

In principle, \texttt{AugPT} is a plug-and-play method compatible with prompt tuning frameworks that perform internal knowledge distillation from unlabeled images. However, to the best of our knowledge, in the field of prompt tuning, PromptKD remains the only reproducible approach with the setting of \textit{internal knowledge distillation + unlabeled images}, which limits the extent of direct empirical comparison.

To further evaluate the adaptability of \texttt{AugPT}, we include a comparison with the UPL~\cite{huang2022upl} framework, which shares a similar design philosophy. UPL is a prompt tuning method that also relies solely on unlabeled images, using the frozen pretrained CLIP model as supervision and guiding CoOp-based prompt tuning through the query results of images. For a fair comparison, when integrating \texttt{AugPT} into this setup, we similarly replace the teacher model and Filtering Gate with the foundation CLIP model. In addition, we follow the original UPL implementation, where model performance is measured by fine-tuning accuracy using the full label set as base classes. To prevent data leakage, test images are excluded from tuning process.

As in Tab.~\ref{tabS10}, integrating the \texttt{AugPT} strategy further improves the performance of UPL, highlighting the adaptability of \texttt{AugPT} to other distillation-based prompt tuning frameworks.

\begin{table}[t]
    \centering
    \small

\setlength\tabcolsep{4pt}

\begin{tabular}{c|ccc|cc} 
\toprule
\rowcolor{gray!10} \textbf{Method} & \begin{tabular}[c]{@{}>{\cellcolor{gray!10}}c@{}}\textbf{Learnable}\\\textbf{Params}\end{tabular} & \begin{tabular}[c]{@{}>{\cellcolor{gray!10}}c@{}}\textbf{Memory }\\\textbf{(MB)}\end{tabular} & \begin{tabular}[c]{@{}>{\cellcolor{gray!10}}c@{}}\textbf{Tuning }\\\textbf{FPS}\end{tabular} & \begin{tabular}[c]{@{}>{\cellcolor{gray!10}}c@{}}\textbf{HM }\\\textbf{Acc.}\end{tabular} & \textbf{$\Delta$}  \\ 
\midrule
PromptKD                                          & 1.01 M                                                                                                           & 3319.1                                                                                                       & 44.3                                                                                                        & 83.35                                                                                                    &                    \\
\rowcolor{cyan!15} \textbf{AugPT}      & 1.01 M                                                                                                           & 4808.7                                                                                                       & 36.1                                                                                                        & 84.33                                                                                                    & +0.98              \\
\bottomrule
\end{tabular}
    \caption{Fine-tuning computation cost of PromptKD backbone and \texttt{AugPT} on StanfordCars. }
    \label{tabS9}

\end{table}

\begin{figure}[t]
  \centering
  \includegraphics[width=0.9\linewidth]{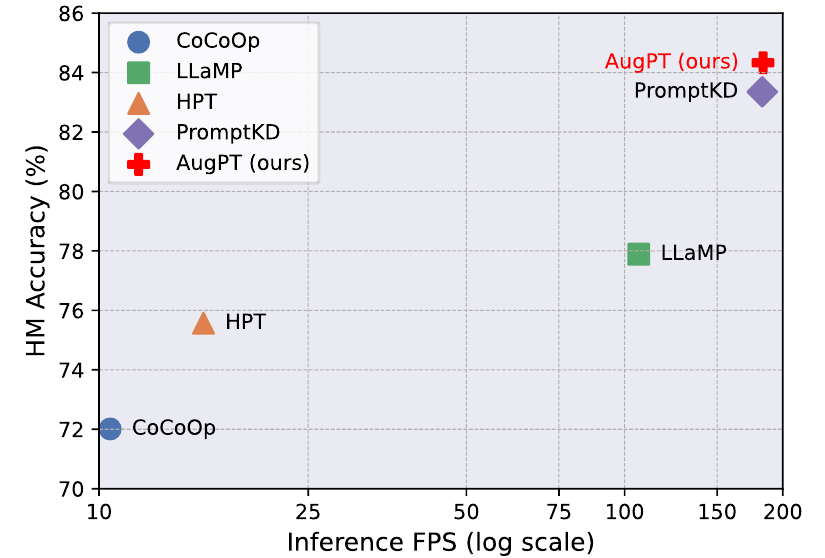}
  \caption{Inference Frames Per Second (FPS) and HM accuracy of \texttt{AugPT} and other data-driven baselines on StanfordCars. Larger FPS value indicates faster inference speed. }
  \label{Figure 6}
\end{figure}

\subsection{Computational Cost} \label{sec:B5}

\paragraph{Fine-Tuning Cost on Base Tasks.}
In Tab.~\ref{tabS9}, we evaluate the computational overhead of \texttt{AugPT} compared to the PromptKD backbone during base-class fine-tuning. Using the StanfordCars \cite{krause2013cars} dataset as an example, we report metrics including the number of learnable parameters, GPU memory usage, image processing speed measured by Frames Per Second (FPS) during fine-tuning, and HM performance for comparison.

We observe that \texttt{AugPT} possesses moderate increases in GPU memory usage and processing time, which can be attributed to the additional computational cost introduced by parallel-handled augmented views. Nonetheless, since \texttt{AugPT} does not attach more learnable parameters and avoids the overhead of collecting external knowledge, the overall increase in fine-tuning overhead remains relatively low. Moreover, since \texttt{AugPT} is not influenced by query-time bottleneck which is unavoidable in LLM-based prompt tuning methods, its fine-tuning FPS (36.1) has a significant advantage over the typical LLM-guided models (normally $<$10). Overall, compared to other data-driven prompt tuning approaches, the fine-tuning cost of \texttt{AugPT} is acceptable.

\paragraph{Inference Cost on New Tasks.}
For inference-stage computational cost, we compare FPS as the evaluation metric in Fig.~\ref{Figure 6}, benchmarking \texttt{AugPT} against advanced data-driven prompt tuning models including HPT \cite{wang2024hpt}, LLaMP \cite{zheng2024llamp}, and PromptKD \cite{li2024promptkd}. It is evident that \texttt{AugPT} achieves faster inference speeds than LLM-guided models and basically remains consist with its backbone, PromptKD, while realizing higher HM performance. This demonstrates the parameter efficiency of \texttt{AugPT} during inference.

\section{Broader Impacts} \label{sec:C}

Our proposed \texttt{AugPT} represents a further extension of research in CLIP-based prompt tuning. As the foundational CLIP model is widely deployed in real-world applications, \texttt{AugPT} offers practical social value by enabling faster adaptation of pre-trained CLIP models to specific downstream tasks or data distributions. Moreover, since \texttt{AugPT} requires no additional external knowledge or data compared to its backbone, it is particularly valuable for prompt tuning in data-scarce scenarios, which are commonly existed in the real world.

At present, we have not identified any ethical concerns or negative societal impacts arising from this research. Nevertheless, we intend to continuously evaluate and monitor the real-world deployment of this research to prevent any potential misuse or unintended applications.

\section{Limitations and Future Work} \label{sec:D}
Although \texttt{AugPT} surpasses current SOTA method using only internal augmentation without introducing external knowledge or additional learnable parameters, we believe there is still room for improvement. Firstly, as mentioned in Sec.~\ref{sec:B6}, the design of \texttt{AugPT} is tightly coupled with distillation-based prompt tuning, making it less adaptable to other prompt tuning paradigms. Secondly, following most existing studies \cite{zhou2022coop, khattak2023promptsrc, li2025dpc, guo2025mmrl} in this area, \texttt{AugPT} focuses on improving performance for recognition-related tasks, while its applicability to other format of tasks (e.g., semantic segmentation) remains unexplored. Finally, the parallel processing of augmented views introduces a moderate increase in computational cost. In future work, we plan to extend \texttt{AugPT} to support a broader range of backbone models and downstream tasks, and consider improving its computational efficiency.